\documentclass{article}

\PassOptionsToPackage{numbers, compress}{natbib}

\usepackage[main, final]{neurips_2026}

\usepackage[utf8]{inputenc} 
\usepackage[T1]{fontenc}    
\usepackage{hyperref}       
\usepackage{url}            
\usepackage{booktabs}       
\usepackage{amsfonts}       
\usepackage{nicefrac}       
\usepackage{microtype}      
\usepackage{xcolor}         
\usepackage{amsmath}
\usepackage{graphicx}
\usepackage{multirow}
\usepackage{subcaption}
\usepackage[table]{xcolor}

\title{SteerCast: Retrieval-Based Latent Steering for Decoder-Only Time Series Forecasting}

\author{%
  Van Dai Do, Huu Hiep Nguyen, Minh Hoang Nguyen, Hung Le \\
  Deakin's Applied Artificial Intelligence Initiative\\
  Deakin University, Geelong, Australia\\
  \texttt{v.do@deakin.edu.au} \\
}

\begin{document}

\maketitle

\begin{abstract}
Time series forecasting aims to predict future values from historical observations and auxiliary features. We propose \textbf{SteerCast}, a retrieval-based latent steering method that improves decoder-only forecaster at inference time, without updating its parameters. SteerCast constructs a database from the training set by storing a representation of each history window together with a \emph{steering vector} computed in the forecaster's latent space, defined as the difference between representations induced by the ground-truth continuation and by the model's own prediction. At test time, SteerCast retrieves nearest neighbors for a query history, aggregates their steering vectors, and injects the resulting signal into the forecaster's hidden states at every step of autoregressive generation, guiding predictions toward trajectories consistent with similar training cases. Experiments across diverse multivariate benchmarks and multiple horizons show that SteerCast consistently improves forecasting accuracy over the fine-tuned backbone and retrieval-based baselines, while requiring no additional training beyond the original fine-tuning and using only the training set as a retrieval corpus.
\end{abstract}

\section{Introduction}
Accurate time series forecasting is a requirement in many real-world systems, including energy management \citep{deb2017review, 8489399}, finance \citep{DBLP:journals/financial_time_series} , economics \citep{franses1998time, RePEc:ucp:jnlbus:v:39:y:1965:p:139} , and healthcare \citep{kaushik2020ai, zhang2024trajectory}. Traditional approaches, such as ARIMA \citep{math10213988}, rely on explicit statistical assumptions to capture temporal dependencies, but they often struggle with the high-dimensional, non-linear patterns found in modern datasets.

Transformers \citep{vaswani2017attention} have become a dominant backbone for long-term time series forecasting, motivating a large body of work that adapts attention-based architectures to improve efficiency and accuracy on long sequences. Representative examples include sparse and efficient attention for long inputs \citep{zhou2021informer}, decomposition-based architectures that separate trend and seasonal components \citep{wu2021autoformer,zhou2022fedformer}, and alternative representations that better exploit periodic structure \citep{wutimesnet}. More recently, patch-based tokenization has improved scalability and generalization by turning long histories into compact token sequences, as demonstrated by \citet{Yuqietal-2023-PatchTST}. Separately, iTransformer \citep{liu2023itransformer} models multivariate structure through variate-centric tokens, enabling effective cross-variable interactions.

The decoder-only time-series models inherit the advantages of next-token generation, flexible context lengths, and strong transfer across domains, and they are increasingly used as high-capacity backbones for forecasting \citep{das2024decoderonlyfoundationmodeltimeseries,liu2025timerxl,shi2025timemoe}. However, even after fine-tuning, forecasting errors often persist in specific regimes, such as rare or atypical patterns, abrupt distribution shifts and interruptions, and horizon-dependent failure modes driven by cumulative error propagation over long horizons \citep{chen2025conceptdrift,hyndman2025fits,lazcano2025multihorizon}. Improving a fine-tuned forecaster typically requires additional training, ensembling, or further scaling, which increases compute and deployment cost \citep{zanotti2025ensemblingcost}. It can also reduce robustness by over-specializing to a target dataset or evaluation condition under distribution shift \citep{wortsman2022wisefinetuning}. This paper explores an alternative route: improving an \emph{already fine-tuned} decoder-only forecaster at inference time, without updating its parameters. Our approach is inspired by steering techniques, which are broader trend in language modeling to improve generalization \citep{liu2024context}.   While input-space retrieval augmentation has been widely studied~\citep{han2025raft,liu2024retrievalaugmented}, directly prepending retrieved histories is impractical for decoder-only time series foundation models: their context windows are strict and self-attention's $\mathcal{O}(L^2)$ cost grows as $(k{+}1)^2$ when $k$ neighbors are prepended~\citep{das2024decoderonlyfoundationmodeltimeseries,shi2025timemoe,vaswani2017attention}.

We propose \textbf{SteerCast}, a retrieval-based inference-time method that improves fine-tuned decoder-only time series forecasters via latent-space steering. SteerCast constructs a database from the training set, where each entry stores (i) a retrieval key that represents the historical context and (ii) a \emph{steering vector} defined in the latent space of the forecaster. At test time, SteerCast retrieves training histories most similar to the current input window, aggregates their steering vectors, and injects the resulting signal into the forecaster's hidden states during generation. The key design choice is that steering vectors are computed from the forecaster's own internal representations, capturing how the latent trajectory changes when conditioning on the ground-truth future versus the model's initial prediction. This yields a model-aware correction signal that transfers forecast-relevant adjustments from similar training cases while preserving the backbone's learned capability. To keep interventions stable, we apply a simple similarity-based modulation and normalization when injecting the steering vector.

We evaluate SteerCast on ten multivariate forecasting benchmarks spanning diverse domains, frequencies, and sequence lengths, and report MSE and MAE across multiple horizons. Across three state-of-the-art decoder-only time series forecasters, Time-MoE \citep{shi2025timemoe}, Timer-XL \citep{liu2025timerxl} and TimesFM \citep{das2024decoderonlyfoundationmodeltimeseries}, SteerCast consistently improves over the corresponding fine-tuned models, showing that retrieval-guided latent steering is an effective and practical mechanism for inference-time refinement.

\paragraph{Contributions.}
Our main contributions are:
(i) we introduce SteerCast, a latent steering method that improves decoder-only forecasters at inference time without parameter updates;
(ii) we propose a database construction procedure that stores retrieval keys and latent steering vectors, together with a lightweight injection rule for stable steering during generation; and
(iii) we demonstrate consistent gains across diverse multivariate forecasting datasets, horizons, and modern decoder-only backbones.

\section{Related Work}
\paragraph{Transformer-based Time Series Forecasting.}

Transformers \citep{vaswani2017attention} have evolved into the dominant backbone for time series modeling, broadly bifurcating into encoder-centric and decoder-centric architectures. Encoder-based methods focus on extracting robust representations for direct prediction: Informer \citep{zhou2021informer} pioneered efficient long-sequence modeling via sparse attention, while PatchTST \citep{Yuqietal-2023-PatchTST} significantly enhanced performance by introducing patch-based tokenization and channel independence. Recent encoder innovations include iTransformer \citep{liu2023itransformer}, which employs an inverted structure to better capture multivariate correlations, and TimeXer \citep{wang2024timexer}, which specifically aligns exogenous variables with endogenous series. Conversely, decoder-based architectures frame forecasting as a generative autoregressive task, often drawing inspiration from Large Language Models (LLMs). AutoTimes \citep{liu2024autotimes} demonstrates the adaptability of frozen LLMs for sequential time series generation, while Timer-XL \citep{liu2025timerxl} establishes a native large-scale foundation model using variable-resolution decoding. To further scale model capacity, Time-MoE \citep{shi2025timemoe} incorporates a Mixture-of-Experts (MoE) design to handle diverse temporal dynamics efficiently. While prior methods typically improve forecasting performance by modifying the training pipeline, our method targets models that are already fine-tuned for each downstream dataset and horizon, and improves their predictions purely at inference time without any further training.

\paragraph{Retrieval-based Forecasting.}
Retrieval has long been used in forecasting through nearest-neighbor and exemplar-based strategies that reuse similar historical patterns. Contemporary retrieval-based methods build a database of historical windows and retrieve relevant examples to guide prediction, either by directly aggregating retrieved futures or by combining retrieved information with a parametric predictor. RATD \citep{liu2024retrievalaugmented} integrates a trained retriever into a diffusion-based generative framework; by conditioning on retrieved historical segments, it allows for stochastic sampling that improves both prediction accuracy and uncertainty quantification. Similarly, RAFT \citep{han2025raft} retrieves historical segments with patterns similar to the current input and leverages their subsequent values as a predictive signal, typically as part of a trained forecasting pipeline. Both methods also require retrieved sequences to fit within the backbone's context window, which is restrictive for fixed-budget foundation models. In contrast, SteerCast uses retrieval to select relevant training cases but applies their information as a latent-space correction, without parameter updates or extra input tokens.

\paragraph{Latent Steering and Representation Interventions.}
A separate line of work studies controlling model behavior by intervening on internal representations, often via additive directions in activation space. Early evidence shows that adding a suitable vector to hidden states can steer a frozen language model toward desired generations \citep{subramani-etal-2022-extracting}. More recently, Wiliński et al.~\citep{wilinski2024unveiling_tsalw} show that time-series foundation models can be steered by intervening on internal activations, and demonstrate simple synthetic edits such as introducing sinusoidal structure into an initially constant signal. In LLM literature, in-context vectors recast in-context learning as a latent steering direction computed from demonstration representations and applied at inference time \citep{liu2024context}. SteerCast adapts this latent intervention perspective to time series forecasting: rather than steering outputs by modifying inputs or retraining parameters, we compute steering vectors from the forecaster's own latent trajectories (ground-truth continuation versus model-predicted continuation) and inject these vectors during generation to refine forecasts.

\begin{figure*}[tbp]
\begin{center}
    \includegraphics[width=\linewidth]{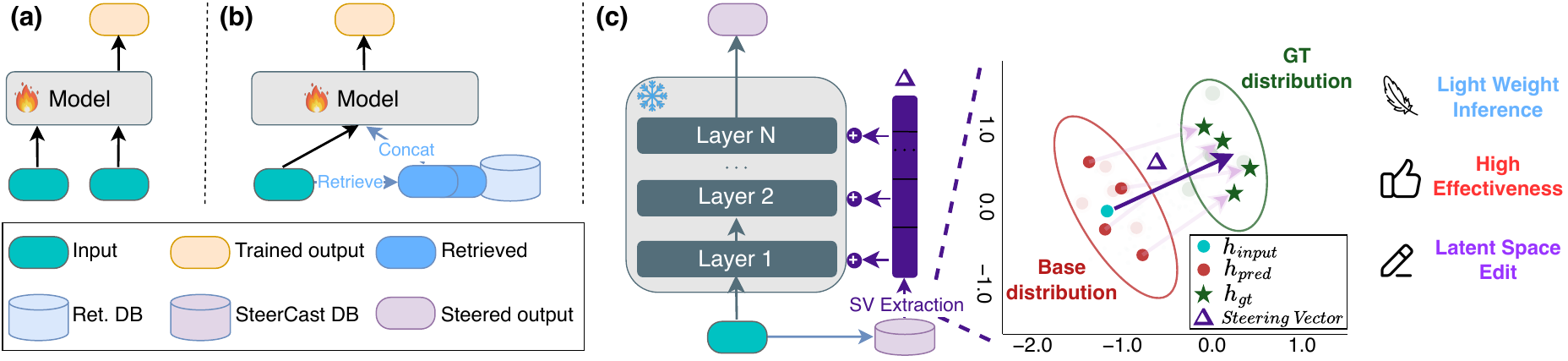}
\end{center}
\caption{Schematic illustrations for improving decoder-based time-series forecasters, including \textbf{(a)} model fine-tuning \textbf{(b)} retrieval-based fine-tuning and \textbf{(c)} our approach \textbf{SteerCast}. SteerCast investigates and demonstrates how to effectively improve forecasting performance with latent space editing while requiring no gradient updates where well-developed methods are not capable of.
}
\label{fig:illustation}  
\end{figure*}

\section{Method}
\subsection{Overview}

\paragraph{Problem Formulation.}Let \(S \in \mathbb{R}^{C \times T}\) be a multivariate time series with \(C\) channels and length \(T\).
Given an input \(\mathbf{X} =[\mathbf{x}_1, \dots, \mathbf{x}_C] \in \mathbb{R}^{C \times P}\), the forecasting task is to predict the future \(\mathbf{Y} = [\mathbf{y}_1,\dots,\mathbf{y}_C] \in \mathbb{R}^{C \times F}\), where \(P\) is the look-back window length and \(F\) is the forecasting horizon. When $\mathcal{F}$ uses patch-based tokenization with patch length $p$, the look-back window of $P$ raw time steps is mapped to $N_P = \lceil P/p \rceil$ input tokens, and the horizon of $F$ raw time steps is produced as $N_F = \lceil F/p \rceil$ forecast tokens. Throughout, we reserve $P, F$ for raw time steps and use $N_P, N_F$ whenever indexing hidden states, retrieval keys, or autoregressive generation steps; for non-patched backbones $p=1$ so $N_P = P$ and $N_F = F$. Following the channel-independent formulation \citep{liu2023itransformer}, we decompose the multivariate series into univariate sequences and process each channel independently using a shared backbone \(\mathcal{F}\).

\begin{figure*}[tbp]
\begin{center}
    \includegraphics[width=\linewidth]{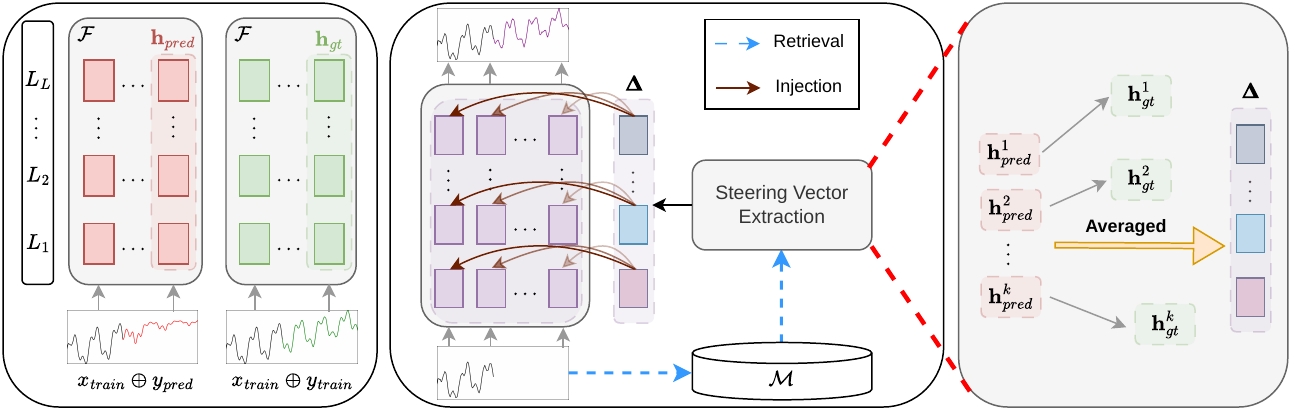}
\end{center}
\caption{\textbf{SteerCast overview.} \textbf{(Left)} For each training example , we run the fine-tuned forecaster \(\mathcal{F}\) to obtain an initial prediction $\textbf{h}_{pred}$ and compute representations for the concatenated sequences \([\mathbf{x_{train}};\mathbf{y_{pred}}]\) and \([\mathbf{x_{train}};\mathbf{y_{train}}]\); their difference yields a steering vector. \textbf{(Middle)} During autoregressive forecasting, we add \(\boldsymbol{\Delta}\) into the hidden states at each generation step at every generation step to guide the forecast toward trajectories consistent with retrieved training cases. \textbf{(Right)} We retrieve the top-\(k\) training histories whose keys are closest to the current query, then average their steering vectors to obtain \(\boldsymbol{\Delta}\).}
\label{fig1:steercast}  
\end{figure*}

\paragraph{Latent Forecast Residuals.}A fine-tuned decoder-only forecaster \(\mathcal{F}\) makes structured errors: similar histories elicit similar forecast deviations, and errors compound over long horizons even after extensive training. Retraining \(\mathcal{F}\) is expensive and can hurt robustness under distribution shift. We instead exploit this structure through SteerCast (Figure~\ref{fig1:steercast}), which steers \(\mathcal{F}\)'s latent states at inference time. We first run \(\mathcal{F}\) on the training set to build a memory \(\mathcal{M}\) of key-value pairs \((\mathbf{r}, \boldsymbol{\Delta})\): for each univariate training sample \((\mathbf{x}, \mathbf{y}) \in \mathcal{D}_{\mathrm{train}}\), \(\mathbf{r}\) is a representation of the historical context (used as the \emph{retrieval key}), and \(\boldsymbol{\Delta}\) is a \emph{steering vector} in the latent space of \(\mathcal{F}\) (used as the value). At test time, given a query series, we compute its key \(\mathbf{r}_q\), retrieve the top-\(k\) most similar keys in \(\mathcal{M}\), and aggregate their steering vectors to obtain \(\boldsymbol{\Delta}\). We then inject \(\boldsymbol{\Delta}\) into the hidden states of \(\mathcal{F}\) during autoregressive generation at every step via a simple linear update, thereby augmenting the fine-tuned forecaster with forecast-relevant latent adjustments distilled from similar training examples and improving prediction quality on new sequences.

\subsection{Database Construction}
\paragraph{Retrieval Key (\(\mathbf{r}\)).} The key should represent histories that elicit similar forecasts close and unrelated histories far apart. Since this similarity depends on how $\mathcal{F}$ processes a window rather than its surface statistics, we derive the key from $\mathcal{F}$ itself. Let $\mathbf{Z}(\mathbf{x}) \in \mathbb{R}^{N_P \times d}$ denote the final-layer hidden states for input $\mathbf{x}$, with $d$ the hidden dimension. To capture the look-back window's global context as the query anchor at inference, we average $\mathbf{Z}$ over the temporal dimension to obtain the key $\mathbf{r}$:
\begin{equation}
\mathbf{r}(\mathbf{x}) = \frac{1}{N_P} \sum_{t=1}^{N_P} \mathbf{Z}_t(\mathbf{x}) \in \mathbb{R}^d.
\end{equation}

\paragraph{Steering Vector (\(\boldsymbol{\Delta}\)).} For every training pair \((\mathbf{x}_{\mathrm{train}}, \mathbf{y}_{\mathrm{train}}) \in \mathcal{D}_{\mathrm{train}}\), SteerCast forms two continuations of the same history: a \emph{source} \([\mathbf{x}_{\mathrm{train}};\mathbf{y}_{\mathrm{pred}}]\) with \(\mathbf{y}_{\mathrm{pred}} = \mathcal{F}(\mathbf{x}_{\mathrm{train}})\), and a \emph{target} \([\mathbf{x}_{\mathrm{train}};\mathbf{y}_{\mathrm{train}}]\), where \(\mathbf{y}_{\mathrm{pred}}\) and \(\mathbf{y}_{\mathrm{train}}\) have equal length. Let \(\mathbf{h}(\mathbf{s}) = [h_1, \dots, h_L] \in \mathbb{R}^{L \times d}\) denote the concatenation of last-token residual-stream states of \(\mathcal{F}\) at each block's output. The steering vector is the difference of \(\mathbf{h}\) under the two continuations:
\begin{equation}
\boldsymbol{\Delta} = \mathbf{h}\bigl([\mathbf{x}_{\mathrm{train}};\mathbf{y}_{\mathrm{train}}]\bigr) - \mathbf{h}\bigl([\mathbf{x}_{\mathrm{train}};\mathbf{y}_{\mathrm{pred}}]\bigr) = \mathbf{h}_{\mathrm{gt}} - \mathbf{h}_{\mathrm{pred}}.
\end{equation}

Intuitively, \(\boldsymbol{\Delta}\) captures the latent displacement from the model's base predictive trajectory (source) to the trajectory consistent with the ground truth (target). Since \(\boldsymbol{\Delta}\) is formed by contrasting representations induced by the ground-truth continuation and the model-generated continuation under the same history \(\mathbf{x}_{\mathrm{train}}\), it cancels representational content that depends only on the history and isolates the displacement attributable to the choice of continuation. Although read off at the last-token position, under causal masking, this is the only token whose receptive field spans the full continuation, so $\boldsymbol{\Delta}$ summarizes the cumulative model-vs-truth disagreement over the entire rollout rather than a horizon-end-only signal; the per-step strength of the intervention is then handled by the cosine gate in Eq.~\eqref{eq:gating}. Storing these steering vectors enables inference-time correction via lightweight additive steering, avoiding any gradient updates. Finally, SteerCast stores \(\mathbf{r}(\mathbf{x}_{\mathrm{train}})\) as the database key for retrieval and associates it with \(\boldsymbol{\Delta}\) as the corresponding steering value, which is later injected at inference time to refine forecasts.                

\subsection{Inference Time Steering}
At test time, we estimate the latent forecast residual at the query by 
nonparametric retrieval and inject the estimate into the forecaster's 
autoregressive rollout. Let \(\mathbf{x} \in \mathbb{R}^{1 \times P}\) 
denote the univariate input history window. We first compute its 
retrieval key \(\mathbf{r}^q = \mathbf{r}(\mathbf{x})\) and measure 
proximity to each stored entry 
\(m^j = (\mathbf{r}^j, \boldsymbol{\Delta}^j) \in M\) under Euclidean 
distance: \(d^j \;=\; \|\mathbf{r}^q - \mathbf{r}^j\|_2, m^j \in M.\label{eq:distance}\) We then select the \(k\) entries with smallest \(d^j\) and aggregate 
their stored residuals by mean pooling:
\begin{equation}
\boldsymbol{\Delta}(\mathbf{x}) \;=\; \frac{1}{k}\sum_{j=1}^{k}
\boldsymbol{\Delta}^j.
\label{eq:aggregation}
\end{equation}

Given a forecasting horizon of $F$ raw time steps, the backbone autoregressively generates $N_F = \lceil F/p \rceil$ forecast tokens. Let $\mathbf{h}_{f,l} \in \mathbb{R}^{1 \times d}$ denote the residual-stream state at block $l \in \{1, \dots, L\}$'s output when generating the $f$-th forecast token, $f \in \{1, \dots, N_F\}$, and let $\boldsymbol{\Delta}^{(l)}(\mathbf{x}) \in \mathbb{R}^{1 \times d}$ be the layer-$l$ segment of the aggregated residual. A naive 
implementation would simply add 
\(\boldsymbol{\Delta}^{(l)}(\mathbf{x})\) to \(\mathbf{h}_{f,l}\) at 
every step. This has two failure modes during autoregressive 
rollout: a constant additive shift continues to push the latent state after the original gap has closed, and the magnitude of the shift 
can vary substantially across queries, occasionally driving the 
trajectory off the manifold the backbone was trained on. We therefore 
control both the \emph{direction} and the \emph{magnitude} of the 
intervention.

\paragraph{Direction: When to steer.} We introduce a non-negative cosine 
gate that weakens the update once the current hidden state is already 
aligned with the steering direction:
\begin{equation}
\alpha_{f,l} \;=\; b + \mathrm{ReLU}\!\Bigl(-\cos\bigl(\mathbf{h}_{f,l}, 
\boldsymbol{\Delta}^{(l)}(\mathbf{x})\bigr) + m\Bigr)^{p},
\label{eq:gating}
\end{equation}
with fixed hyperparameters \((b, m, p)\). The gate has three regimes 
that follow directly from the ReLU. When 
\(\cos(\mathbf{h}_{f,l}, \boldsymbol{\Delta}^{(l)}) \ge m\) — the 
hidden state is already pointed in the steering direction beyond margin 
\(m\) — the gate collapses to its floor \(b\), preventing 
over-correction. When \(\cos < m\), the gate exceeds \(b\) by an amount 
that grows with misalignment, and the exponent \(p\) controls how 
sharply this growth concentrates near the margin (larger \(p\) gives a 
more selective gate). The floor \(b > 0\) ensures a small baseline 
correction even in the aligned regime, which we found necessary to 
prevent the steering from disengaging entirely once the trajectory 
briefly drifts toward the right direction.

\paragraph{Magnitude: How strongly to steer.} We \(\ell_2\)-normalize each 
layer segment of the aggregated residual before applying the gated 
update:
\begin{equation}
\widetilde{\mathbf{h}}_{f,l}
\;=\;
\mathbf{h}_{f,l}
+
\lambda \,\alpha_{f,l}\,
\frac{\boldsymbol{\Delta}^{(l)}(\mathbf{x})}
{\|\boldsymbol{\Delta}^{(l)}(\mathbf{x})\|_2 + \epsilon},
\label{eq:steering-update}
\end{equation}
where \(\lambda\) is a global steering strength and \(\epsilon\) a numerical-stability constant. Normalization separates the \emph{direction} of the residual (which retrieval has estimated) from its \emph{scale} (which retrieval has not), replacing the latter with a single tunable \(\lambda\). The resulting per-step, per-layer perturbation magnitude is bounded by \(\lambda\,\alpha_{\max}\)--where \(\alpha_{\max} = b + (1+m)^{p}\) is the maximum value of the cosine gate in Eq.~\eqref{eq:gating}--regardless of the data or backbone, and \(\epsilon\) prevents division blow-up when \(\|\boldsymbol{\Delta}^{(l)}\|_2 \approx 0\).

\section{Experiments} \label{sec:experiments}
\subsection{Experimental Settings}
\paragraph{Datasets.} We benchmark SteerCast on ten datasets spanning diverse variates, lengths, and frequencies. ETT (four subsets) \citep{zhou2021informer} contains electricity transformer measurements at 15-minute intervals; Exchange \citep{lai2018lstnet} tracks daily exchange rates for eight countries; Weather \citep{weather_dataset} contains 21 German weather indicators; and Illness \citep{illness_dataset} reports weekly influenza-like illness ratios. From the Monash archive \citep{godahewa2021monash}, we use three univariate series: US Births, SaugeenDay (Canadian river discharge), and Sunspots.

\paragraph{Backbones and Baselines.} We use Time-MoE \citep{shi2025timemoe}, Timer-XL \citep{liu2025timerxl} and TimesFM \citep{das2024decoderonlyfoundationmodeltimeseries} as backbones, which are state-of-the-art decoder-only Transformers for time series forecasting. We compare against three baselines: \textbf{FT}, the simple method that fine-tunes model on the downstream dataset; \textbf{RAFT} \citep{han2025raft}, a retrieval-augmented method that retrieves top-\(k\) similar historical patches and uses their subsequent segments alongside the input to produce the forecast; and \textbf{RAF} \citep{tire2024retrieval}, a training-free method that prepends retrieved (history, future) examples as in-context demonstrations to the backbone.

\begin{table*}[t]
\caption{MSE comparison of SteerCast (SC), Fine-tune (FT), RAF \citep{tire2024retrieval}, and RAFT \citep{han2025raft} across 10 datasets, averaged over all prediction lengths, under the \textbf{various look-back window} setting. \textit{Promotion}: \% MSE reduction of SC over RAFT. Full results in Appendix~\ref{appendix:full_results}.}
\label{tab:various_to_all_transposed}
\centering
\scriptsize
\setlength{\tabcolsep}{3pt}
\renewcommand{\arraystretch}{0.75}
\resizebox{\textwidth}{!}{%
\begin{tabular}{cl cccccccccc}
\toprule
& \textbf{Method} & ETTh1 & ETTh2 & ETTm1 & ETTm2 & Exch. & Wthr. & Ill. & Births & Saug. & Suns. \\
\midrule
\multirow{5}{*}{\rotatebox{90}{Time-MoE}}
& FT          & 0.389 & 0.366 & 0.350 & 0.367 & 0.428 & 0.241 & 3.392 & 0.855 & 0.999 & 0.429 \\
& RAF         & 0.398 & 0.356 & 0.358 & 0.373 & 0.440 & 0.249 & 3.322 & 0.923 & 0.996 & 0.438 \\
& RAFT        & 0.389 & 0.359 & 0.349 & 0.349 & 0.449 & 0.242 & 3.279 & 0.724 & 0.966 & 0.422 \\
& \textbf{SC (ours)}   & 0.380 & 0.351 & 0.345 & 0.339 & 0.418 & 0.242 & 3.232 & 0.639 & 0.983 & 0.420 \\
& \textit{Promotion}
& \cellcolor{green!20}2.3\%
& \cellcolor{green!20}2.2\%
& \cellcolor{green!20}1.1\%
& \cellcolor{green!20}2.8\%
& \cellcolor{green!20}6.9\%
& \cellcolor{green!20}0.0\%
& \cellcolor{green!20}1.4\%
& \cellcolor{green!20}11.7\%
& -1.8\%
& \cellcolor{green!20}0.5\% \\
\midrule
\multirow{5}{*}{\rotatebox{90}{Timer-XL}}
& FT          & 0.681 & 0.377 & 0.384 & 0.279 & 0.546 & 0.252 & 2.894 & 0.748 & 1.082 & 0.399 \\
& RAF         & 0.697 & 0.424 & 0.391 & 0.289 & 0.561 & 0.262 & 2.941 & 0.674 & 1.118 & 0.395 \\
& RAFT        & 0.709 & 0.414 & 0.403 & 0.304 & 0.530 & 0.266 & 2.893 & 0.623 & 1.033 & 0.391 \\
& \textbf{SC (ours)}   & 0.681 & 0.372 & 0.381 & 0.280 & 0.518 & 0.255 & 2.889 & 0.625 & 1.029 & 0.391 \\
& \textit{Promotion}
& \cellcolor{green!20}3.9\%
& \cellcolor{green!20}10.1\%
& \cellcolor{green!20}5.5\%
& \cellcolor{green!20}7.9\%
& \cellcolor{green!20}2.3\%
& \cellcolor{green!20}4.1\%
& \cellcolor{green!20}0.1\%
& -0.3\%
& \cellcolor{green!20}0.4\%
& \cellcolor{green!20}0.0\% \\
\midrule
\multirow{5}{*}{\rotatebox{90}{TimesFM}}
& FT                 & 0.468 & 0.421 & 0.444 & 0.310 & 0.503 & 0.270 & 3.213 & 0.667 & 1.003 & 0.181 \\
& RAF                & 0.462 & 0.424 & 0.437 & 0.325 & 0.499 & 0.268 & 3.156 & 0.687 & 1.037 & 0.371 \\
& RAFT               & 0.452 & 0.424 & 0.440 & 0.307 & 0.498 & 0.264 & 3.167 & 0.648 & 0.985 & 0.173 \\
& \textbf{SC (ours)} & 0.452 & 0.409 & 0.434 & 0.302 & 0.485 & 0.255 & 3.083 & 0.538 & 0.956 & 0.171 \\
& \textit{Promotion} & \cellcolor{green!20}0.0\% & \cellcolor{green!20}3.6\% & \cellcolor{green!20}1.5\% & \cellcolor{green!20}1.4\% & \cellcolor{green!20}2.6\% & \cellcolor{green!20}3.4\% & \cellcolor{green!20}2.7\% & \cellcolor{green!20}16.9\% & \cellcolor{green!20}2.9\% & \cellcolor{green!20}1.2\% \\
\bottomrule
\end{tabular}%
}
\end{table*}

\begin{table*}[t]
\caption{MSE comparison of SteerCast (SC), Fine-tune (FT), RAF \citep{tire2024retrieval}, and RAFT \citep{han2025raft} across 10 datasets, averaged over all prediction lengths, under the \textbf{fixed look-back window} setting. \textit{Promotion}: \% MSE reduction of SC over RAFT. Full results in Appendix~\ref{appendix:full_results}.}
\label{tab:512_to_all}
\centering
\scriptsize
\setlength{\tabcolsep}{3pt}
\renewcommand{\arraystretch}{0.75}
\resizebox{\textwidth}{!}{%
\begin{tabular}{cl cccccccccc}
\toprule
& \textbf{Method} & ETTh1 & ETTh2 & ETTm1 & ETTm2 & Exch. & Wthr. & Ill. & Births & Saug. & Suns. \\
\midrule
\multirow{5}{*}{\rotatebox{90}{Time-MoE}}
& FT          & 0.407 & 0.537 & 0.401 & 0.509 & 0.425 & 0.237 & 3.838 & 1.012 & 1.150 & 0.566 \\
& RAF         & 0.400 & 0.513 & 0.394 & 0.518 & 0.431 & 0.242 & 3.848 & 0.953 & 1.139 & 0.541 \\
& RAFT        & 0.402 & 0.518 & 0.391 & 0.519 & 0.435 & 0.236 & 3.868 & 0.851 & 1.106 & 0.500 \\
& \textbf{SC (ours)}   & 0.394 & 0.488 & 0.387 & 0.513 & 0.420 & 0.238 & 3.817 & 0.832 & 1.124 & 0.483 \\
& \textit{Promotion}
& \cellcolor{green!20}2.0\%
& \cellcolor{green!20}5.8\%
& \cellcolor{green!20}1.0\%
& \cellcolor{green!20}1.2\%
& \cellcolor{green!20}3.4\%
& -0.8\%
& \cellcolor{green!20}1.3\%
& \cellcolor{green!20}2.2\%
& -1.6\%
& \cellcolor{green!20}3.4\% \\
\midrule
\multirow{5}{*}{\rotatebox{90}{Timer-XL}}
& FT          & 0.641 & 0.394 & 0.420 & 0.294 & 0.537 & 0.248 & 2.524 & 0.704 & 1.202 & 0.426 \\
& RAF         & 0.653 & 0.400 & 0.425 & 0.305 & 0.457 & 0.265 & 2.606 & 0.629 & 1.112 & 0.410 \\
& RAFT        & 0.661 & 0.405 & 0.422 & 0.309 & 0.461 & 0.271 & 2.584 & 0.608 & 1.095 & 0.413 \\
& \textbf{SC (ours)}    & 0.624 & 0.386 & 0.418 & 0.292 & 0.436 & 0.246 & 2.528 & 0.620 & 1.102 & 0.402 \\
& \textit{Promotion}
& \cellcolor{green!20}5.6\%
& \cellcolor{green!20}4.7\%
& \cellcolor{green!20}0.9\%
& \cellcolor{green!20}5.5\%
& \cellcolor{green!20}5.4\%
& \cellcolor{green!20}9.2\%
& \cellcolor{green!20}2.2\%
& -2.0\%
& -0.6\%
& \cellcolor{green!20}2.7\% \\
\midrule
\multirow{5}{*}{\rotatebox{90}{TimesFM}}
& FT          & 0.502 & 0.399 & 0.418 & 0.303 & 0.428 & 0.223 & 3.300 & 0.268 & 1.015 & 0.338 \\
& RAF         & 0.521 & 0.383 & 0.416 & 0.296 & 0.420 & 0.221 & 3.200 & 0.550 & 1.040 & 0.323 \\
& RAFT        & 0.508 & 0.384 & 0.412 & 0.299 & 0.422 & 0.219 & 3.250 & 0.256 & 0.984 & 0.325 \\
& \textbf{SC (ours)}   & 0.476 & 0.381 & 0.394 & 0.291 & 0.408 & 0.213 & 3.163 & 0.249 & 0.952 & 0.319 \\
& \textit{Promotion}
& \cellcolor{green!20}6.3\%
& \cellcolor{green!20}0.9\%
& \cellcolor{green!20}4.4\%
& \cellcolor{green!20}2.7\%
& \cellcolor{green!20}3.3\%
& \cellcolor{green!20}2.6\%
& \cellcolor{green!20}2.7\%
& \cellcolor{green!20}2.8\%
& \cellcolor{green!20}3.3\%
& \cellcolor{green!20}1.8\% \\
\bottomrule
\end{tabular}%
}
\end{table*}

\paragraph{Evaluation Protocols.} For each dataset, we (i) construct the retrieval database $\mathcal{M}$ from the chronological training split only, (ii) select hyperparameters such as the number ofretrieved neighbors $k$ on the held-out validation split, following~\citep{han2025raft}, and (iii) evaluate on the held-out test split (the chronologically last $20\%$ of the series). The validation block sits between the training database and the test set in time and is excluded from $\mathcal{M}$, so no database entry spans timesteps that any test window predicts; we further verify this in Section~\ref{leakage_ablation}. We report mean squared error (MSE) and mean absolute error (MAE), and consider forecasting horizons $F \in \{96, 192, 336, 720\}$ except for Illness, where $F \in \{24, 36, 48, 60\}$.
All evaluations are conducted in the multivariate setting, using all channels of each dataset.

\subsection{Results}
\subsubsection{Various Look-back Window Forecasting}
\paragraph{Setup.} We vary the look-back window length for different forecasting horizons. Specifically, for horizons \(\{96,192,336,720\}\), we use input lengths \(\{512,1024,2048,3072\}\), respectively. This setting builds a separate retrieval database for each input--prediction length pair.

\paragraph{Results.} Table~\ref{tab:various_to_all_transposed} shows that \textbf{SteerCast performs consistently across all three forecasting backbones}. It achieves the lowest or tied-lowest MSE on \textbf{25 of 30} dataset--backbone entries. Compared with RAF, SteerCast reduces MSE by \textbf{7.4\%} on average, with gains observed across most entries. These results show that horizon-specific retrieval databases provide effective steering signals across diverse datasets and backbones.

\subsubsection{Fixed Look-back Window Forecasting}
\paragraph{Setup.} We next consider a more efficient setting where the input length is fixed to \(512\) for all prediction horizons \(\{96,192,336,720\}\). We build only the retrieval database for the \(512\text{-}96\) setting and reuse the same retrieval keys and steering vectors across all horizons. This evaluates whether SteerCast can transfer a single retrieval-guided correction across different forecasting lengths.

\paragraph{Results.} Table~\ref{tab:512_to_all} shows that \textbf{SteerCast remains effective under the fixed look-back setting}. Across 30 dataset--backbone entries, SteerCast achieves the lowest or tied-lowest MSE on \textbf{24} entries. On average, SteerCast reduces MSE by \textbf{5.1\%} over FT, \textbf{5.3\%} over RAF, and \textbf{2.7\%} over RAFT. These results suggest that a single database built from the \(512\text{-}96\) setting can still provide transferable steering signals across longer horizons, reducing the need for horizon-specific database construction.

\section{Ablation Studies and Model Analysis}
\begin{table}[t]
\centering
\small
\caption{Robustness analyses on ETT with Time-MoE, reporting MSE averaged over prediction horizons. \textbf{Left:} Temporal-isolation ablation: default vs.\ strict variant dropping the last $P+F=608$ train candidates ($k{=}1$, $\lambda{=}0.01$, Euclidean retrieval). \textbf{Right:} Distribution shift, using only the first 50\% of training data. $\Delta$: relative MSE reduction of SteerCast (SC) over RAFT. MAE in appendix.}
\label{tab:robustness}

\begin{minipage}[t]{0.38\linewidth}
\centering
\setlength{\tabcolsep}{6pt}
\begin{tabular}{l cc}
\toprule
\textbf{Protocol} & ETTh1 & ETTh2 \\
\midrule
Default      & 0.342 & 0.290 \\
Strict       & 0.343 & 0.287 \\
\bottomrule
\end{tabular}
\subcaption{Temporal-isolation ablation.}
\label{tab:leakage_ablation}
\end{minipage}%
\hfill
\begin{minipage}[t]{0.60\linewidth}
\centering
\setlength{\tabcolsep}{5pt}
\begin{tabular}{l cccc c}
\toprule
\textbf{Dataset} & \textbf{FT} & \textbf{RAF} & \textbf{RAFT} & \textbf{SC} & $\boldsymbol{\Delta}$ \\
\midrule
ETTh1 & 0.441 & 0.421 & 0.409 & \textbf{0.406} & 0.9\% \\
ETTh2 & 0.630 & 0.531 & 0.529 & \textbf{0.515} & 2.7\% \\
\bottomrule
\end{tabular}
\subcaption{Temporal distribution shift (first 50\% of train).}
\label{tab:temporal_shift}
\end{minipage}
\end{table}

\subsection{Temporal Isolation Ablation} \label{leakage_ablation}                      
To verify that SteerCast's gains are not driven by residual boundary effects between the database and the test split, we rebuild the database under a stricter temporal-isolation protocol. The default protocol is already leak-free by construction: the database is built only from the chronological training portion of each series, with the entire validation block acting as a natural buffer. The stricter variant additionally drops the last $P + F = 608$ candidate windows on the train side at the $512\!\rightarrow\!96$ setting, enlarging this buffer further. Table~\ref{tab:leakage_ablation} compares the two protocols on ETTh1 and ETTh2 with Time-MoE: the strict variant changes MSE by about $0.3\%$ on ETTh1 and $1.0\%$ on ETTh2, with the two deltas pointing in opposite directions (slightly higher error on ETTh1, slightly lower on ETTh2). If boundary windows were carrying forward-looking information into the database, removing them should consistently hurt the strict protocol; instead, the change is small in magnitude and inconsistent in sign, indicating that boundary entries are not a meaningful source of SteerCast's improvement.

\subsection{SteerCast Works Well Under Temporal Distribution Shift}
We further evaluate SteerCast under a temporally shifted setting by restricting the available training data for all methods to the first 50\% of the training set. Since the ETT datasets are chronologically ordered, this creates a harder evaluation scenario in which the training memory is more temporally separated from the held-out test period. As shown in Table~\ref{tab:temporal_shift}, SteerCast remains robust under this setting, achieving the best overall performance on both ETTh1 and ETTh2. Compared with FT and retrieval-based baselines, SteerCast consistently obtains lower average error, indicating that the retrieved latent residuals remain useful even when the available memory is limited to earlier temporal regimes. These results support SteerCast as an effective inference-time correction mechanism under temporal distribution shift.

\subsection{Database Design}

\begin{figure*}[t]
\centering
\caption{\textbf{Database design ablations.} \textbf{(Left)} Database protocol on Time-MoE, comparing a single reused database with per-horizon databases. \textbf{(Right)} ETTh2 with Time-MoE: forecasting MSE versus retrieval database size (10\%--100\%), with Fine-tune and RAFT as horizontal baselines.}
\includegraphics[width=0.8\textwidth]{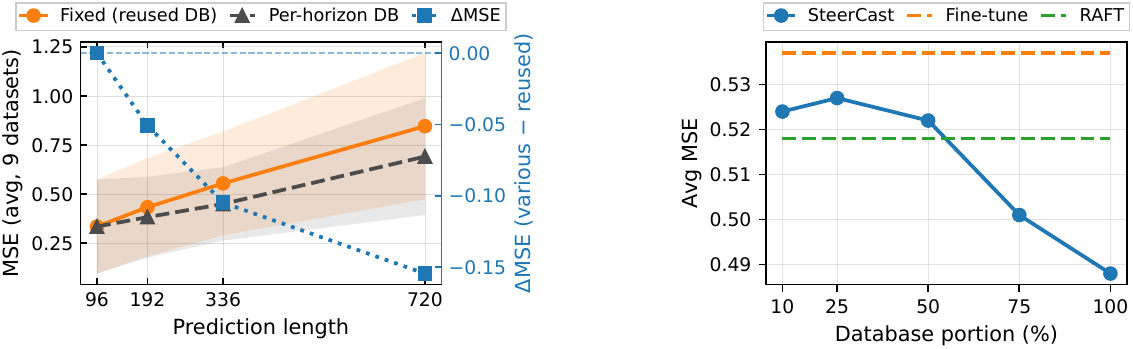}
\label{fig:database_combined}
\end{figure*}

\paragraph{Fixed vs.\ various-history protocol.} We compare the \emph{fixed-history} and \emph{various-history} database building protocols. Figure~\ref{fig:database_combined} (Left) reports MSE averaged over nine datasets (excluding Illness due to its different forecast lengths), with a mean$\pm$std band showing cross-dataset variability. Fixed-history is already a strong default, but various-history becomes increasingly beneficial as the horizon grows: the gap between the two curves widens, and $\Delta$MSE (per-horizon minus reused) becomes more negative, indicating that horizon-specific memory better captures long-range dynamics.

\paragraph{Database Size.} We study the effect of database size on ETTh2, using Time-MoE as the forecaster. The retrieval databases are constructed using 10\%, 25\%, 50\%, 75\%, and 100\% of the training set and report results in Figure~\ref{fig:database_combined} (Right). As the database grows, SteerCast's average MSE decreases from 0.524 (10\%) to 0.488 (100\%), suggesting that larger pools more reliably retrieve close historical matches and thus yield more effective correction vectors. SteerCast consistently outperforms the Fine-tune baseline across all sizes, and it surpasses RAFT once the database reaches 50\% (and above), with the largest gains at 75\%--100\%. The mild non-monotonicity at 10\%--50\% indicates that smaller pools can still retrieve regime-mismatched neighbors, whereas a sufficiently large memory reduces this risk and improves the consistency of steering.

\begin{table}[t]
\centering
\caption{\textbf{Steering Analysis}. \textbf{(a)} insertion position of \(\boldsymbol{\Delta}\) on ETTh1, averaged over horizons \(\{96,192,336,720\}\) with history 512, weight 0, neighbor 1, and \(\lambda=0.01\). \textbf{(b)} gating ablation, where cosine-based gating is disabled and the gate is reduced to a constant \(\alpha_{f,l}\equiv b\).}
\label{tab:steercast_ablation_compact}
\small
\setlength{\tabcolsep}{4.5pt}

\begin{minipage}[t]{0.45\linewidth}
\centering
\textbf{(a) \(\boldsymbol{\Delta}\) insertion position}

\vspace{3pt}
\begin{tabular}{lcc}
\toprule
\textbf{Layer} & \textbf{Avg MSE \(\downarrow\)} & \textbf{Avg MAE \(\downarrow\)} \\
\midrule
All layers   & \textbf{0.394} & \textbf{0.416} \\
First layer  & 0.403 & 0.426 \\
Middle layer & 0.406 & 0.429 \\
Last layer   & 0.402 & 0.426 \\
\bottomrule
\end{tabular}
\end{minipage}
\hfill
\begin{minipage}[t]{0.50\linewidth}
\centering
\textbf{(b) Gating ablation}

\vspace{2pt}
\begin{tabular}{lccc}
\toprule
\textbf{Dataset} & \textbf{Pred.} & \textbf{MSE} \(\downarrow\) & \textbf{MAE} \(\downarrow\) \\
\midrule
\multirow{2}{*}{ETTh1} 
& 720  & 0.500 & 0.508 \\
& Average & 0.410 & 0.429 \\
\midrule
\multirow{2}{*}{ETTh2} 
& 720  & 0.920 & 0.663 \\
& Average & 0.527 & 0.480 \\
\bottomrule
\end{tabular}
\end{minipage}

\end{table}

\subsection{Anatomy of the Steering Rule}
\paragraph{Layer Impact of Steering Vector.} We ablate the depth at which \(\boldsymbol{\Delta}\) is applied, comparing insertion at a single transformer block (first, middle, or last) against all blocks, with retrieval fixed on ETTh1 at history length 512. Table~\ref{tab:steercast_ablation_compact} (a) shows that all-layer insertion performs best, while single-layer steering raises average MSE by \(2.0\%\)--\(3.0\%\). This indicates that \(\boldsymbol{\Delta}\)'s correction signal is not localized to one depth and is most effective when distributed across the forecaster.

\paragraph{Gating ablation.} To isolate the benefit of adaptive gating, we disable the ReLU term in Equation~\eqref{eq:gating}, thus the gate becomes a constant \(\alpha_{f,l}\equiv b\). Table~\ref{tab:steercast_ablation_compact} (b) reports the resulting errors on ETTh1 and ETTh2 with the same retrieval and steering settings. Without the gate, performance degrades as the horizon grows, with particularly large long-horizon errors at 720 steps, suggesting that a uniform, ungated intervention is more prone to accumulated mis-calibration during autoregressive rollout.

\begin{figure*}[t]
\centering
\caption{\textbf{Steering Analysis.} \textbf{(Left)} Sensitivity to steering strength $\lambda$, where smaller values perform best. \textbf{(Right)} ETTh1 retrieval-key layer ablation: first, middle, or last layer for the retrieval key.}
\includegraphics[width=0.8\textwidth]{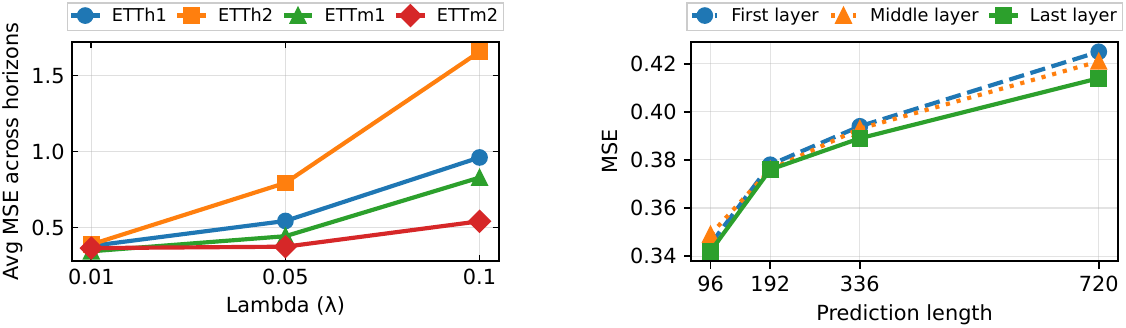}
\label{fig:steering_combined}
\end{figure*}

\paragraph{Sensitivity to Steering Strength $\lambda$.} Across all four ETT datasets (Figure~\ref{fig:steering_combined}, Left), a small $\lambda=0.01$ consistently achieves the lowest MSE and MAE, while $\lambda=0.05$ and $0.1$ substantially degrade performance. Overly strong latent intervention over-corrects the backbone dynamics and pushes trajectories off-manifold, with the effect amplifying at longer horizons. We use $\lambda=0.01$ as the default and recommend tuning within a narrow range around this value.

\paragraph{Retrieval Key.} We compare the results when selecting different layers for the retrieval representation, using ETTh1 with Time-MoE. We evaluate three choices: using the first, middle, or last layer to form the retrieval key (our default is the last layer). As reported in Figure~\ref{fig:steering_combined} (Right), the last-layer representation yields the lowest average error and remains consistently strong across horizons, while the first and middle layers are slightly worse, particularly at longer horizons. This suggests that higher-layer representations better capture task-relevant temporal patterns for retrieval, leading to more compatible neighbors and a more reliable transferred steering signal.

\subsection{Efficiency}
Table~\ref{tab:cache_storage} shows that the two database protocols require nearly identical cache storage, with minor variation across datasets. Inference-time comparison is in Table~\ref{tab:runtime}. SteerCast adds minimal overhead over the FT baseline, comparable to RAF and substantially less than RAFT. We provide more results for storage and latency in Appendices \ref{app:storage_scale} and \ref{app:build_time}; and time-complexity analysis in Appendix~\ref{app:complexity}.

\begin{table}[t]
\centering
\begin{minipage}[t]{0.48\textwidth}
\centering
\small
\setlength{\tabcolsep}{4pt}
\caption{Cache storage (GB) of SteerCast using Time-MoE under different database protocols.}
\label{tab:cache_storage}
\begin{tabular}{@{}lcccc@{}}
\toprule
\textbf{Protocol} & ETTh1 & ETTh2 & ETTm1 & \textbf{Avg.} \\
\midrule
Fixed   & 4.494 & 4.957 & 5.822 & \textbf{5.091} \\
Various & 4.884 & 4.900 & 5.929 & \textbf{5.238} \\
\bottomrule
\end{tabular}
\end{minipage}
\hfill
\begin{minipage}[t]{0.48\textwidth}
\centering
\small
\setlength{\tabcolsep}{5pt}
\caption{Runtime comparison (s/iter and relative to Base) using Time-MoE as the forecaster.}
\label{tab:runtime}
\begin{tabular}{@{}lcccc@{}}
\toprule
\textbf{Method} & FT & SteerCast & RAF & RAFT \\
\midrule
s/iter & 0.0921 & 0.1034 & 0.1089 & 0.1473 \\
Rel.   & 1.00$\times$ & 1.12$\times$ & 1.18$\times$ & 1.60$\times$ \\
\bottomrule
\end{tabular}
\end{minipage}
\end{table}

\subsection{Additional Analysis.} We report further ablations on (i) the choice of retrieval metric, (ii) retrieval key choice, (iii) gating ablations, (iv) the number of retrieved neighbors, (v) retrieval quality and failure mode, (vi) additional comparison with output-level ensembling, (vii) distance-weighted neighbor ablation in Appendices \ref{app:retrieval_metric}, \ref{app:retrieval_key}, \ref{app:gate_ablation}, \ref{app:k_finegrained}, \ref{app:retrieval_quality}, \ref{app:ensemble_baseline}, \ref{app:weighted_agg}. Overall, these results are consistent with our main findings and further support the robustness of our method across retrieval configurations.

\section{Conclusion}
We introduce SteerCast, a test-time latent steering method for decoder-only time series forecasting. Unlike standard retrieval-augmentation, SteerCast injects a model-aware correction—derived from historical representation errors—directly into the hidden states during autoregressive generation. Benchmarks across ten datasets and state-of-the-art backbones demonstrate that SteerCast consistently improves accuracy without requiring additional training, particularly for long-horizon forecasting where it mitigates error accumulation.

\section*{Acknowledgments}
This research was funded (partially or fully) by the Australian Government through the Australian Research Council.
Dr Hung Le is the recipient of an Australian Research Council Discovery Early Career Researcher Award (project number DE250100355) funded by the Australian Government.

\bibliography{example}
\bibliographystyle{plainnat}



\newpage
\appendix
\section{Appendix for SteerCast.}
\subsection{Qualitative Analysis} 
\begin{figure*}[h]
\centering
\caption{Qualitative retrieval relevance on ETTh1 (512$\rightarrow$96) with Time-MoE as the forecaster. \textbf{(Left)} The inference query and its nearest retrieved training sequence closely overlap over the look-back region, indicating high retrieval fidelity. \textbf{(Right)} The retrieved training future provides a coherent continuation that matches the target-horizon dynamics, while the base model prediction exhibits noticeable drift, motivating the use of the retrieved example for steering at inference time.}
\includegraphics[width=0.9\linewidth]{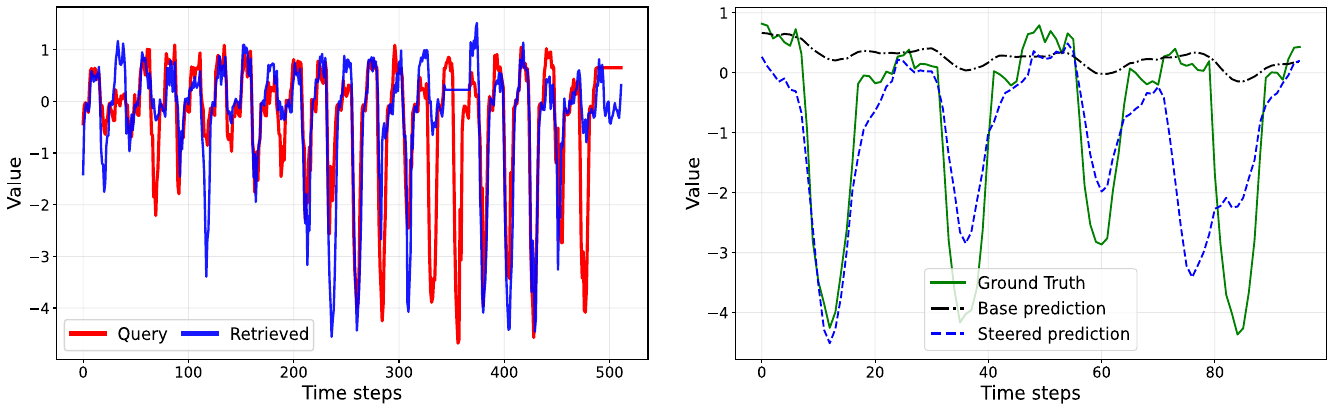}
\label{fig:retrieval}
\end{figure*}

Figure~\ref{fig:retrieval} illustrates why SteerCast improves forecasting at inference time. \textbf{(Left)} The retrieved training example closely matches the query over the look-back window, indicating that the retrieval key identifies genuinely similar historical contexts. This alignment makes the transferred correction meaningful: a well-matched neighbor supplies a compatible future pattern from which SteerCast derives an informative adjustment. \textbf{(Right)} Compared to the base forecast, which drifts as the horizon progresses, the steered prediction tracks the ground truth more closely, indicating that the intervention corrects the trajectory rather than merely smoothing the output. Together, these panels link retrieval relevance to downstream gains: SteerCast leverages similar examples to construct a model-aware correction that nudges generation toward a plausible continuation while largely preserving the backbone's dynamics.

\subsection{Different Retrieval Metric} \label{app:retrieval_metric}

\begin{figure}[t]
\centering
\caption{Average MSE across US\_Births, Saugeenday and Sunspot datasets for SteerCast using Cosine vs.\ Euclidean neighbor retrieval at forecasting horizons $\{96,192,336,720\}$, using Time-MoE as the forecaster. Euclidean retrieval yields consistently lower error, with the gap widening at longer horizons. Lower MSE is better.}
\includegraphics[width=0.6\linewidth]{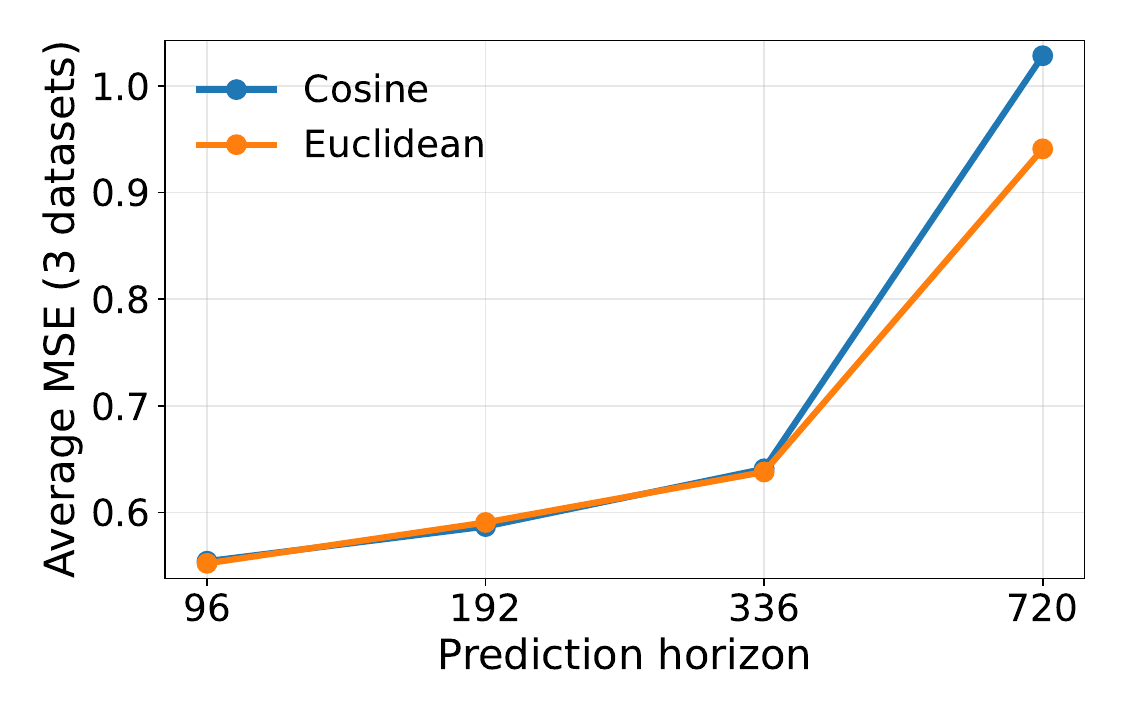}
\label{fig:different_retrieval}
\end{figure}

In this section, we replace Euclidean distance with Cosine distance for nearest-neighbor retrieval and report the comparison in Figure~\ref{fig:different_retrieval}. After averaging MSE across the three datasets at each horizon, Euclidean retrieval is consistently comparable or better than cosine, and the gap widens as the horizon increases: performance is similar at 96/192, while Euclidean is clearly stronger at 336/720. One plausible explanation is that cosine similarity is invariant to vector magnitude, whereas Euclidean distance retains scale information that can matter for selecting neighbors whose latent correction $\boldsymbol{\Delta}$ is appropriately calibrated. In autoregressive forecasting, small calibration errors can compound over time, so retrieving a neighbor with a more compatible representation scale can yield more stable long-horizon corrections. In addition, cosine similarity can become less informative when representations exhibit strong anisotropy (i.e., many vectors concentrate in a narrow cone), causing many candidates to have similar cosine scores and reducing effective neighbor discrimination.

\subsection{Retrieval-key Ablation} \label{app:retrieval_key}
In this section, we study the effect of different ways to represent the keys in our database. Mean-pooling, our default choice for the retrieval key $\textbf{}r$, outperforms the last-token alternative at every horizon we evaluate. Table~\ref{tab:ablation_retkey} reports the comparison on ETTh1 with all other hyperparameters held fixed. Substituting the last input token's final-layer hidden state for the mean-pooled key worsens MSE by $+2.03\%$ at $h{=}96$, $+4.96\%$ at $h{=}192$, $+0.48\%$ at $h{=}336$, and $+0.60\%$ at $h{=}720$, for an average penalty of $+1.83\%$ in MSE (and a similar trend in MAE). One might expect the last-token key to be preferable a priori because, under causal masking, it is the only position whose hidden state has a receptive field over the entire lookback. We find that this intuition does not translate into improved retrieval: in practice, mean-pooling acts as a temporal regularizer that aggregates evidence from multiple subsequences within the input window and reduces sensitivity to the precise boundary state, and this benefit empirically dominates the receptive-field advantage of the last token. We therefore retain mean-pooling as the default; \emph{the last-token variant is uniformly worse on this benchmark.}

\begin{table}[t]                                      
\centering
\caption{Retrieval-key formulation ablation on ETTh1 (Time-MoE, lookback 512, $k{=}1$). \textbf{Mean-pooling (our default) outperforms the last-token alternative at every horizon.} The rightmost column reports the MSE penalty incurred by
switching from mean-pooling to the last-token key; all four entries are positive, indicating that the last-token variant is uniformly worse.}    
\label{tab:ablation_retkey}
\small                            
\begin{tabular}{lcccc}                  
\toprule                                
& \multicolumn{2}{c}{Mean pooling (default)} & \multicolumn{2}{c}{Last token} \\
\cmidrule(lr){2-3}\cmidrule(lr){4-5}       
Horizon & MSE & MAE & MSE & MAE \\  
\midrule               
96  & \textbf{0.345} & \textbf{0.376} & 0.352 & 0.389 \\ 
192 & \textbf{0.383} & \textbf{0.402} & 0.402 & 0.427 \\ 
336 & \textbf{0.414} & \textbf{0.430} & 0.416 & 0.431 \\ 
720 & \textbf{0.500} & \textbf{0.508} & 0.503 & 0.510 \\ 
\midrule                 
Average & \textbf{0.411} & \textbf{0.429} & 0.418 & 0.439 \\                   
\midrule               
\multicolumn{1}{l}{Last-token MSE penalty (avg)} & \multicolumn{4}{c}{$\mathbf{+1.83\%}$ \emph{worse than mean-pooling}} \\
\bottomrule                         
\end{tabular}                                         \end{table}





\begin{table}[t]
\centering
\setlength{\tabcolsep}{6pt}
\caption{Sensitivity to gate hyperparameters $(b, m, p)$. Sweep on Exchange with Time-MoE, varying one parameter at a time and holding the other two at the default. MSE/MAE are averaged over
horizons $\{96, 192, 336, 720\}$. The default $(0.1, 0.1, 1.25)$ is shaded; \textbf{bold} marks the best value of each metric.}
\label{tab:gate_sensitivity}
\begin{tabular}{ccc cc}
\toprule
$b$ & $m$ & $p$ & MSE $\downarrow$ & MAE $\downarrow$ \\
\midrule
\rowcolor{gray!15} 0.1 & 0.1 & 1.25 & 0.4200 & 0.4510 \\
\midrule
\multicolumn{5}{l}{\emph{Sweep $b$ (baseline gate)}} \\
0.0 & 0.1 & 1.25 & 0.4206 & 0.4511 \\
0.5 & 0.1 & 1.25 & 0.4215 & 0.4521 \\
\midrule
\multicolumn{5}{l}{\emph{Sweep $m$ (alignment margin)}} \\
0.1 & 0.0 & 1.25 & 0.4178 & 0.4513 \\
0.1 & 0.5 & 1.25 & 0.4209 & 0.4521 \\
\midrule
\multicolumn{5}{l}{\emph{Sweep $p$ (gate sharpness)}} \\
0.1 & 0.1 & 1.00 & 0.4214 & 0.4517 \\
0.1 & 0.1 & 1.50 & 0.4203 & 0.4510 \\
\bottomrule
\end{tabular}
\end{table}

\subsection{Gate Hyperparameters Ablation}
\label{app:gate_ablation}
The cosine gate in Equation~\eqref{eq:gating} introduces three fixed hyperparameters, set to $(b, m, p) = (0.1, 0.1, 1.25)$ in all main experiments. To check whether SteerCast is sensitive to these choices, we sweep each one in isolation on Exchange with Time-MoE, holding the other two at the default and averaging over horizons $\{96, 192, 336, 720\}$ (Table~\ref{tab:gate_sensitivity}). The first observation is that the spread across all seven configurations is small---between $0.4178$ and $0.4215$ MSE, a relative range of under $1\%$---so SteerCast is robust to moderate perturbations of the gate. Within this narrow band, the trends are interpretable. Increasing $p$ from $1.0$ to $1.5$ slightly improves accuracy, consistent with $p$ controlling how sharply the gate concentrates near the misalignment margin: a more selective gate engages strongly only when the hidden state is genuinely misaligned with $\boldsymbol{\Delta}^{(l)}$ and otherwise stays near its floor. Increasing the baseline floor $b$ from $0.1$ to $0.5$ degrades performance, indicating that a heavier baseline correction continues to perturb the latent state even when no correction is needed; setting $b{=}0$ recovers comparable accuracy but removes the small floor that keeps steering engaged through transient alignment dips. Increasing the margin $m$ from $0$ to $0.5$ also degrades performance, because a larger margin causes the gate to fire even on moderately aligned states and over-corrects predictions that are already on track. The configuration $(0.1, 0.0, 1.25)$ slightly improves over the default on this dataset; we retain $m{=}0.1$ as the default because the small positive margin acts as a buffer against numerical noise in the cosine score and the difference is well within the $1\%$ band observed across the sweep.

\subsection{Fine-grained Sensitivity to the Number of Neighbors}
\label{app:k_finegrained}

To verify that SteerCast's performance is not driven by aggressive tuning of $k$, we sweep $k \in \{1,2,3,4,5,6,8,10,12,16,20\}$ at fixed look-back length $512$ on five datasets with Time-MoE and report the average MSE in Table~\ref{tab:k_finegrained}. The behavior is structured rather than brittle: ETTh1 favors small $k$, ETTm1 is largely insensitive, ETTh2 improves to a moderate $k$ and then plateaus, and the more irregular Monash datasets (US-Births, SaugeenDay) benefit from larger $k$. This pattern is consistent with $k$ trading off local correction against noise-reducing averaging---a property of the retrieval pool rather than a hyperparameter to be aggressively tuned. Tuning the number of retrieved neighbors is also standard in retrieval-augmented forecasting~\citep{han2025raft}.

\begin{table}[t]
\centering
\caption{Fine-grained $k$ sensitivity (average MSE across horizons $\{96,192,336,720\}$, history $512$, Time-MoE).}
\label{tab:k_finegrained}
\small
\setlength{\tabcolsep}{4pt}
\begin{tabular}{l ccccccccccc}
\toprule
\textbf{Dataset} & 1 & 2 & 3 & 4 & 5 & 6 & 8 & 10 & 12 & 16 & 20 \\
\midrule
ETTh1      & \textbf{0.394} & 0.397 & 0.398 & 0.399 & 0.400 & 0.402 & 0.405 & 0.410 & 0.407 & 0.405 & 0.404 \\
ETTh2      & 0.488 & 0.490 & 0.474 & \textbf{0.470} & 0.476 & 0.474 & 0.473 & 0.473 & 0.473 & 0.476 & 0.476 \\
ETTm1      & \textbf{0.387} & \textbf{0.387} & \textbf{0.387} & 0.389 & 0.391 & 0.391 & 0.391 & 0.396 & 0.392 & 0.392 & 0.392 \\
US-Births  & 0.941 & 0.986 & 0.954 & 0.939 & 0.930 & 0.923 & 0.918 & 0.915 & 0.912 & 0.907 & \textbf{0.904} \\
SaugeenDay & 1.135 & 1.143 & 1.139 & 1.139 & 1.137 & 1.137 & 1.136 & 1.130 & 1.134 & 1.132 & \textbf{1.128} \\
\bottomrule
\end{tabular}
\end{table}

\subsection{Retrieval Quality and Failure-mode Analysis}
\label{app:retrieval_quality}

We probe whether SteerCast's residual error is driven by the steering mechanism itself or by the quality of the retrieved neighbors. On ETTh1 and ETTh2, we vary the database from $10\%$ to $100\%$ of the training set at horizon $720$ with history $512$ and report the average retrieval distance (Euclidean in latent space) alongside MSE in Table~\ref{tab:retrieval_quality}. Both axes shift together: enlarging the database from $10\%$ to $100\%$ approximately halves the average retrieval distance and improves MSE by $23$--$33\%$. This indicates that, when SteerCast underperforms, the bottleneck is the absence of close neighbors in the database rather than instability in the steering rule. The cosine gate (Eq.~\eqref{eq:gating}) and per-layer normalization (Eq.~\eqref{eq:steering-update}) are designed to bound the harm in exactly this regime: when the retrieved direction is poorly aligned with the current latent state, the gate attenuates the update; the gating ablation in Table~\ref{tab:steercast_ablation_compact}(b) shows that removing this protection turns SteerCast from better than RAFT into worse on ETTh2 at horizon $720$ ($0.745\to 0.920$ MSE).

\begin{table}[t]
\centering
\caption{Retrieval quality vs.\ forecasting error. Increasing the database size lowers the average distance to retrieved neighbors, which translates into lower MSE.}
\label{tab:retrieval_quality}
\small
\setlength{\tabcolsep}{8pt}
\begin{tabular}{l c c c}
\toprule
\textbf{Dataset} & \textbf{DB Size} & \textbf{Avg.\ Retrieval Distance} & \textbf{MSE} \\
\midrule
ETTh1 & 10\%  & 14.87 & 0.576 \\
ETTh1 & 100\% & 7.43  & 0.441 \\
\midrule
ETTh2 & 10\%  & 28.68 & 1.112 \\
ETTh2 & 100\% & 8.75  & 0.745 \\
\bottomrule
\end{tabular}
\end{table}

\subsection{Comparison with Output-level Ensembling}
\label{app:ensemble_baseline}

A natural alternative to latent steering is to ensemble the fine-tuned forecast with the retrieved future values directly in the output space. We construct this baseline by averaging the fine-tuned model's prediction with the mean of the top-$4$ retrieved future windows, with the interpolation weight $\eta$ tuned on the validation set ($\eta = 0.1$ selected). Table~\ref{tab:ensemble_baseline} compares the two on nine datasets at fixed history $512$. SteerCast outperforms output-level ensembling on every dataset, reducing average MSE from $0.613$ to $0.542$ ($-11.6\%$). This supports our claim that the gain comes from the \emph{model-aware latent correction} (the difference between $\mathbf{h}_{\mathrm{gt}}$ and $\mathbf{h}_{\mathrm{pred}}$) rather than from the retrieved future values themselves.

\begin{table}[t]
\centering
\caption{SteerCast vs.\ output-level ensembling (FT prediction $+$ mean of top-$4$ retrieved futures, $\eta=0.1$). Average MSE across horizons $\{96,192,336,720\}$ at history $512$ with Time-MoE; lower is better.}
\label{tab:ensemble_baseline}
\small
\setlength{\tabcolsep}{6pt}
\begin{tabular}{l c c}
\toprule
\textbf{Dataset} & \textbf{SteerCast} & \textbf{Ensemble} \\
\midrule
ETTh1       & \textbf{0.394} & 0.415 \\
ETTh2       & \textbf{0.488} & 0.574 \\
ETTm1       & \textbf{0.387} & 0.405 \\
ETTm2       & \textbf{0.513} & 0.546 \\
Exchange    & \textbf{0.420} & 0.487 \\
SaugeenDay  & \textbf{1.124} & 1.310 \\
Sunspots    & \textbf{0.483} & 0.493 \\
US-Births   & \textbf{0.832} & 1.045 \\
Weather     & \textbf{0.238} & 0.241 \\
\midrule
\textbf{Average} & \textbf{0.542} & 0.613 \\
\bottomrule
\end{tabular}
\end{table}

\subsection{Distance-weighted Neighbor Aggregation}
\label{app:weighted_agg}

The default aggregation in Eq.~\eqref{eq:aggregation} is a uniform mean over the top-$k$ steering vectors. A natural alternative is to weight each neighbor by a decreasing function of its retrieval distance, so that closer neighbors contribute more. We tested softmax-weighted and inverse-distance-weighted aggregation in preliminary experiments on the ETT datasets and observed no consistent improvement over the uniform mean (Table~\ref{tab:weighted_agg}): distance-weighted aggregation yields slightly worse MSE and MAE on every ETT subset. A plausible reason is that, since SteerCast already retrieves only the top-$k$ most similar entries (rather than averaging over the full database), most candidates have nearly identical retrieval distances, and exponentiating these small differences amplifies retrieval noise rather than meaningful similarity gradient. We therefore retain uniform mean aggregation for its simplicity, predictability, and one-fewer hyperparameter; richer aggregation schemes (e.g., learned attention over neighbors) are an interesting direction for future work but were not necessary to obtain the reported gains.

\begin{table}[t]
\centering
\caption{Uniform mean (SteerCast) vs.\ distance-weighted neighbor aggregation. Average MSE and MAE across horizons $\{96,192,336,720\}$ at history $512$ with Time-MoE; lower is better.}
\label{tab:weighted_agg}
\small
\setlength{\tabcolsep}{6pt}
\begin{tabular}{l cc cc}
\toprule
& \multicolumn{2}{c}{\textbf{SteerCast}} & \multicolumn{2}{c}{\textbf{Distance-weighted}} \\
\cmidrule(lr){2-3}\cmidrule(lr){4-5}
\textbf{Dataset} & MSE & MAE & MSE & MAE \\
\midrule
ETTh1 & \textbf{0.394} & \textbf{0.416} & 0.401 & 0.420 \\
ETTh2 & \textbf{0.488} & \textbf{0.460} & 0.491 & 0.463 \\
ETTm1 & \textbf{0.387} & \textbf{0.406} & 0.391 & 0.408 \\
ETTm2 & \textbf{0.512} & \textbf{0.449} & 0.515 & 0.451 \\
\bottomrule
\end{tabular}
\end{table}

\subsection{Storage at Scale on Multivariate Datasets}
\label{app:storage_scale}

Under the channel-independent formulation, the steering bank stores one entry per univariate training window, so storage scales with both dataset size and channel count. Table~\ref{tab:storage_scale} reports the cache footprint at history length $512$ on the two highest-dimensional datasets in our benchmark. While the absolute size on Traffic ($862$ channels, $39.12$\,GB) is non-trivial, storage grows substantially \emph{slower than linear} in raw cache entries: Traffic has $\approx 13.7\times$ more entries than Weather, but only $\approx 2.2\times$ larger cache, due to factor-shared layouts in the persistence format. Storage can be further reduced by sub-sampling the database, as shown in Figure~\ref{fig:database_combined} (Right), where SteerCast remains effective at $50$--$75\%$ of the full bank. Approximate-nearest-neighbor indexing and quantized cache entries (e.g., FAISS, product quantization) are natural extensions for very large multivariate corpora.

\begin{table}[t]
\centering
\caption{Cache footprint at history length $512$ on high-dimensional datasets with Time-MoE.}
\label{tab:storage_scale}
\small
\setlength{\tabcolsep}{8pt}
\begin{tabular}{l c c}
\toprule
\textbf{Dataset} & \textbf{Cache (GB)} & \textbf{Cache entries} \\
\midrule
Weather  & 17.54 & 1{,}106{,}616 \\
Traffic  & 39.12 & 15{,}122{,}928 \\
\bottomrule
\end{tabular}
\end{table}

\subsection{Database Construction Time}
\label{app:build_time}

Table~\ref{tab:build_time} reports the average wall-clock time to construct the steering memory cache, averaged over horizons $\{96, 192, 336, 720\}$ with history $512$ on Time-MoE. Construction is performed once offline and amortized across all subsequent inference queries; build time is dominated by the cost of running the forecaster on each training window to extract latent representations.

\begin{table}[t]
\centering
\caption{Average database construction time per dataset (seconds, averaged over horizons $\{96,192,336,720\}$, history $512$, Time-MoE).}
\label{tab:build_time}
\small
\setlength{\tabcolsep}{8pt}
\begin{tabular}{l c}
\toprule
\textbf{Dataset} & \textbf{Time (s)} \\
\midrule
ETTh1 & 14.13 \\
ETTh2 & 14.08 \\
ETTm1 & 16.39 \\
ETTm2 & 16.66 \\
\bottomrule
\end{tabular}
\end{table}

\subsection{Why SteerCast Is Scoped to Decoder-only Forecasters}
\label{app:scope}

SteerCast targets decoder-only autoregressive forecasters because its inference-time intervention is naturally defined on the per-step latent rollout that decoder-only architectures expose. Specifically, the injection in Eq.~\eqref{eq:steering-update} is applied to the residual-stream state $\mathbf{h}_{f,l}$ at each generation step $f \in \{1, \dots, N_F\}$ and each transformer block $l$, which presupposes that the model produces forecast tokens \emph{sequentially} from a left-to-right rollout under causal masking. Encoder-based forecasters (e.g., PatchTST, iTransformer) and direct-prediction models map the entire look-back window to all $N_F$ output tokens in a single forward pass, so there is no per-step latent trajectory to steer; applying SteerCast to such models would require a different formulation of both the stored correction signal (currently the difference between two terminal-token rollouts) and the test-time injection rule (currently per-step gated addition). This is a deliberate scope choice rather than a fundamental limitation: state-of-the-art time series foundation models, including Time-MoE~\citep{shi2025timemoe}, Timer-XL~\citep{liu2025timerxl}, and TimesFM~\citep{das2024decoderonlyfoundationmodeltimeseries}, are decoder-only autoregressive models, and our experiments show consistent improvements across all three. Extending the latent-steering perspective to encoder-based forecasters is an interesting direction for future work.

\subsection{Dataset Details}
\label{app:datasets}

\paragraph{ETT (Electricity Transformer Temperature)} \citep{zhou2021informer} The ETT benchmark consists of four subsets recording load and oil temperature from electricity transformers in two separated counties in China between July 2016 and July 2018. ETTh1 and ETTh2 are sampled hourly, while ETTm1 and ETTm2 are sampled every 15 minutes. Each subset contains 7 variables (6 power-load features and 1 oil temperature target). ETT has become a standard benchmark for long-horizon forecasting and is widely used to evaluate models on multivariate sequences with strong daily and weekly seasonality.

\paragraph{Exchange} \citep{lai2018lstnet} This dataset records the daily exchange rates of eight foreign currencies (Australia, the United Kingdom, Canada, Switzerland, China, Japan, New Zealand, and Singapore) relative to the US dollar from 1990 to 2016. Exchange-rate series are notoriously non-stationary and lack obvious periodic structure, making this a stress test for forecasters that rely on recurring temporal patterns.

\paragraph{Weather} The Weather dataset\footnote{\url{https://www.bgc-jena.mpg.de/wetter/}} contains 21 meteorological indicators (e.g., air temperature, humidity, atmospheric pressure, wind velocity, solar radiation) recorded every 10 minutes throughout 2020 at the Max Planck Institute for Biogeochemistry weather station in Jena, Germany. The high sampling frequency and rich set of co-varying physical signals make it useful for evaluating models on dense, multivariate environmental data.

\paragraph{Illness} The Illness dataset\footnote{\url{https://gis.cdc.gov/grasp/fluview/fluportaldashboard.html}} consists of weekly counts of influenza-like illness (ILI) cases reported to the U.S. Centers for Disease Control and Prevention from 2002 to 2021, expressed as the ratio of ILI patients to total patients seen. The series exhibits strong annual seasonality coupled with substantial year-to-year variability driven by epidemic dynamics. Owing to its low sampling rate, the dataset is comparatively short, which is why we use shorter history and forecast horizons (96/24, 192/36, 256/48, 336/60) rather than the standard $\{96, 192, 336, 720\}$.

\paragraph{Monash univariate series} \citep{godahewa2021monash} The Monash time series forecasting archive aggregates a large collection of single-channel real-world series across diverse domains. We use three subsets that complement ETT and Weather by emphasising domains with distinct temporal structure:

\textbf{US Births.} Daily counts of live births in the United States from 1969 to 1988. The series exhibits both pronounced weekly seasonality (fewer births on weekends) and a slow annual cycle.

\textbf{SaugeenDay.} Daily mean river discharge of the Saugeen River in Ontario, Canada, recorded over several decades. The signal combines smooth seasonal flow patterns with sharp transient peaks driven by precipitation and snowmelt events, making it a useful test of how models handle bursty, partially predictable dynamics.

\textbf{Sunspots.} Monthly counts of observed sunspots, one of the longest-running scientific time series. It exhibits a quasi-periodic $\sim$11-year solar cycle whose amplitude varies substantially across cycles, providing a long-range, slowly-varying signal distinct from the higher-frequency datasets above.

\begin{table}[ht]
\centering
\caption{Summary of dataset statistics. Variates is the number of channels per series; Length is the total number of timesteps; Frequency is the sampling interval.}
\label{tab:dataset_details}
\small
\setlength{\tabcolsep}{6pt}
\begin{tabular}{l c c c l}
\toprule
\textbf{Dataset} & \textbf{Variates} & \textbf{Length} & \textbf{Frequency} & \textbf{Domain} \\
\midrule
ETTh1 / ETTh2 & 7  & 17{,}420 & Hourly      & Energy \\
ETTm1 / ETTm2 & 7  & 69{,}680 & 15-min      & Energy \\
Exchange      & 8  &  7{,}588 & Daily       & Finance \\
Weather       & 21 & 52{,}696 & 10-min      & Climate \\
Illness       & 7  &    966   & Weekly      & Health \\
US Births     & 1  &  7{,}305 & Daily       & Demographics \\
SaugeenDay    & 1  & 23{,}741 & Daily       & Hydrology \\
Sunspots      & 1  & 73{,}931 & Daily       & Astronomy \\
\bottomrule
\end{tabular}
\end{table}

\subsection{Hyperparameters} In this section, we provide the hyperparameters used in our experiments for Time-MoE, Timer-XL and TimesFM models. We note that even though they are identical, we still separate them into two tables for clarification. In addition, we note that these parameters are shared for different forecasting horizons within the same dataset.

\subsection{Shared Parameters}
We provide the universal used parameters values in Table \ref{tab:gating_hparams}.

\begin{table}[t]
\centering
\caption{Fixed gating hyperparameters used in the similarity-based gate in Equation \ref{eq:gating}.}

\begin{tabular}{c l c}
\toprule
\textbf{Symbol} & \textbf{Description} & \textbf{Value} \\
\midrule
$b$ &  Baseline gate & 0.1 \\
$m$ & Margin  &  0.10 \\
$p$ & Power & 1.25 \\
\bottomrule
\end{tabular}
\label{tab:gating_hparams}
\end{table}

\subsubsection{Fixed Look-back Window Forecasting}
We provide hyperparameters for this setting in Table \ref{tab:fixed_lb_hparams}.

\begin{table}[t]
\centering
\caption{Hyperparameters for fixed look-back window forecasting: number of retrieved neighbors ($k$) and steering weight ($\lambda$) for \textsc{Time-MoE}, \textsc{Timer-XL}, and \textsc{TimesFM}. For each model, we keep $k$ constant across all forecasting horizons and history-length variants.}
\small
\setlength{\tabcolsep}{5pt}
\begin{tabular}{l|cc|cc|cc}
\toprule
& \multicolumn{2}{c|}{\textbf{Time-MoE}} & \multicolumn{2}{c|}{\textbf{Timer-XL}} & \multicolumn{2}{c}{\textbf{TimesFM}} \\
\textbf{Dataset} & \textbf{\# Neighbors $k$} & \textbf{$\lambda$} & \textbf{\# Neighbors $k$} & \textbf{$\lambda$} & \textbf{\# Neighbors $k$} & \textbf{$\lambda$} \\
\midrule
ETTh1          & 1  & 0.01 & 1  & 0.01 & 1 & 0.01 \\
ETTh2          & 1  & 0.01 & 16 & 0.01 & 1 & 0.01 \\
ETTm1          & 4  & 0.01 & 16 & 0.01 & 1 & 0.01 \\
ETTm2          & 6  & 0.01 & 4  & 0.01 & 1 & 0.01 \\
exchange\_rate & 1  & 0.01 & 4  & 0.01 & 4 & 0.01 \\
weather        & 1  & 0.01 & 1  & 0.01 & 4 & 0.01 \\
illness        & 1  & 0.01 & 1  & 0.01 & 1 & 0.01 \\
us\_births     & 15 & 0.01 & 2  & 0.01 & 1 & 0.01 \\
saugeenday     & 2  & 0.01 & 13 & 0.01 & 16 & 0.01 \\
sunspots       & 5  & 0.01 & 10 & 0.01 & 1 & 0.01 \\
\bottomrule
\end{tabular}
\label{tab:fixed_lb_hparams}
\end{table}

\subsubsection{Various Look-back Window Forecasting}
We provide hyperparameters for this setting in Table \ref{tab:various_lb_hparams}.

\begin{table}[t]
\centering
\caption{Hyperparameters for various look-back window forecasting: number of retrieved neighbors ($k$) and steering weight ($\lambda$) for \textsc{Time-MoE}, \textsc{Timer-XL}, and \textsc{TimesFM}. For each model, we keep $k$ constant across all forecasting horizons and history-length variants.}
\small
\setlength{\tabcolsep}{5pt}
\begin{tabular}{l|cc|cc|cc}
\toprule
& \multicolumn{2}{c|}{\textbf{Time-MoE}} & \multicolumn{2}{c|}{\textbf{Timer-XL}} & \multicolumn{2}{c}{\textbf{TimesFM}} \\
\textbf{Dataset} & \textbf{\# Neighbors $k$} & \textbf{$\lambda$} & \textbf{\# Neighbors $k$} & \textbf{$\lambda$} & \textbf{\# Neighbors $k$} & \textbf{$\lambda$} \\
\midrule
ETTh1          & 1  & 0.01 & 1  & 0.01 & 1 & 0.01 \\
ETTh2          & 1  & 0.01 & 16 & 0.01 & 4 & 0.01 \\
ETTm1          & 1  & 0.01 & 16 & 0.01 & 1 & 0.01 \\
ETTm2          & 6  & 0.01 & 4  & 0.01 & 1 & 0.01 \\
exchange\_rate & 1  & 0.01 & 4  & 0.01 & 16 & 0.01 \\
weather        & 1  & 0.01 & 1  & 0.01 & 4 & 0.01 \\
illness        & 1  & 0.01 & 1  & 0.01 & 1 & 0.01 \\
us\_births     & 6  & 0.01 & 4  & 0.01 & 4 & 0.01 \\
saugeenday     & 2  & 0.01 & 1  & 0.01 & 1 & 0.01 \\
sunspots       & 4  & 0.01 & 4  & 0.01 & 1 & 0.01 \\
\bottomrule
\end{tabular}
\label{tab:various_lb_hparams}
\end{table}

\subsection{Matched Output-Space Residual Correction}
\label{app:matched-output-residual}

Output-level ensembling with retrieved futures does not directly test whether
retrieved forecast errors are better applied in latent or output space. We
therefore compare SteerCast with an output-residual baseline using the same
retrieval keys, top-$k$ neighbors, and per-dataset $k$. For query $x$, this
baseline predicts
\begin{equation}
\widehat{y}_{\mathrm{out}}(x)
=\widehat{y}(x)+\beta\,\frac{1}{k}
\sum_{j\in\mathcal{N}_k(x)}\bigl(y_j-\widehat{y}_j\bigr),
\label{eq:app-output-residual}
\end{equation}
where $y_j$ and $\widehat{y}_j$ are the ground-truth and model-predicted
continuations of retrieved training example $j$. We use $\beta$ to distinguish
the output correction weight from SteerCast's layer- and step-dependent gate.
We sweep $\beta\in\{0,0.1,0.25,0.5,0.75,1.0\}$ with Time-MoE, history
length $P=512$, and forecast horizon $F=96$.

\begin{table}[htbp]
\centering
\caption{Matched output-residual correction versus latent steering. Entries
are MSE; lower is better. $\beta=0$ recovers the fine-tuned (FT) backbone.}
\label{tab:app-output-residual}
\begin{tabular}{lrrr}
\toprule
Correction & ETTh1 & ETTh2 & ETTm2 \\
\midrule
Output, $\beta=0$ (FT) & 0.345 & 0.300 & \textbf{0.191} \\
Output, $\beta=0.1$ & 0.348 & 0.299 & 0.195 \\
Output, $\beta=0.25$ & 0.371 & 0.310 & 0.207 \\
Output, $\beta=0.5$ & 0.458 & 0.360 & 0.244 \\
Output, $\beta=0.75$ & 0.606 & 0.448 & 0.300 \\
Output, $\beta=1.0$ & 0.815 & 0.576 & 0.376 \\
SteerCast & \textbf{0.342} & \textbf{0.290} & \textbf{0.191} \\
\bottomrule
\end{tabular}
\end{table}

Among the tested output weights, the lowest MSE occurs at $\beta=0$ on
ETTh1 and ETTm2, and at $\beta=0.1$ on ETTh2. SteerCast improves on the
best tested output correction on ETTh1 and ETTh2 and ties it on ETTm2.
Increasing $\beta$ beyond $0.1$ progressively worsens all three results.
Thus, reusing the same neighbors through a scalar-weighted output residual
does not reproduce the benefit of the complete latent-steering procedure.
This comparison supports the proposed method over this particular output
baseline; it does not establish that all output-space correction methods are
inferior or isolate intervention location from normalization and adaptive gating.

\subsection{Steering Pretrained Backbones Without Fine-Tuning}
\label{app:pretrained-steering}

We apply SteerCast directly to pretrained checkpoints without downstream
parameter updates. The memory is constructed from the checkpoint's own
predictions and the training-set ground-truth continuations. We select $k$
on the validation split and fix $\lambda=0.01$, with forecast horizon
$F=96$. The backbone is unfine-tuned in both columns of
Table~\ref{tab:app-pretrained-steering}; the steered variant additionally
uses labeled training examples as its retrieval memory. Consequently, this
is adaptation without parameter updates, rather than a setting with no
access to target-dataset labels.

\begin{table}[htbp]
\centering
\caption{MSE of pretrained checkpoints before and after steering, without
fine-tuning. The steering strength is fixed rather than tuned per dataset.}
\label{tab:app-pretrained-steering}
\begin{tabular}{llrr}
\toprule
Backbone & Dataset & Pretrained & Pretrained + SteerCast \\
\midrule
Time-MoE & ETTh1 & 0.358 & \textbf{0.357} \\
Time-MoE & ETTh2 & 0.302 & \textbf{0.289} \\
Time-MoE & ETTm2 & 0.197 & \textbf{0.195} \\
Time-MoE & Weather & \textbf{0.159} & 0.161 \\
TimesFM & ETTh1 & 0.396 & \textbf{0.391} \\
TimesFM & ETTh2 & 0.326 & \textbf{0.316} \\
\bottomrule
\end{tabular}
\end{table}

Steering reduces MSE in five of the six settings. The largest relative
reductions occur on ETTh2: approximately $4.3\%$ for Time-MoE and $3.1\%$
for TimesFM. Weather instead regresses by approximately $1.3\%$, showing
that the fixed intervention is not uniformly beneficial. These results
establish that prior fine-tuning is not a prerequisite for improvement in
the evaluated settings. Although $\lambda=0$ recovers the unsteered
checkpoint, including this option in validation-based selection would not
guarantee non-degradation on the test set.

\subsection{Temporal Coverage and Retrieval Stress Tests}
\label{app:retrieval-stress}

\paragraph{Full versus reduced training coverage.}
We compare full training coverage with access to only the first $50\%$ of
the chronological training split. The setting uses Time-MoE, $P=512$, and
MSE averaged over $F\in\{96,192,336,720\}$. The retrieval configurations
use $k=1$. Training-data access is reduced for all adapted methods, so this
experiment changes both the available fine-tuning data and retrieval data;
it is not an isolated intervention on memory size. The zero-shot checkpoint
uses neither and is unchanged across coverage levels.

\begin{table}[htbp]
\centering
\caption{MSE under full and first-half training coverage. Relative change
is $100(\mathrm{MSE}_{50\%}/\mathrm{MSE}_{100\%}-1)$, computed from the
reported rounded MSE values.}
\label{tab:app-training-coverage}
\begin{tabular}{llrrr}
\toprule
Dataset & Method & Full & First $50\%$ & Change \\
\midrule
ETTh1 & Zero-shot & 0.445 & 0.445 & $0.0\%$ \\
 & FT & 0.407 & 0.441 & $+8.4\%$ \\
 & RAF & 0.400 & 0.421 & $+5.3\%$ \\
 & RAFT & 0.402 & 0.409 & $+1.7\%$ \\
 & SteerCast & \textbf{0.394} & \textbf{0.406} & $+3.0\%$ \\
\midrule
ETTh2 & Zero-shot & 0.565 & 0.565 & $0.0\%$ \\
 & FT & 0.537 & 0.630 & $+17.3\%$ \\
 & RAF & 0.513 & 0.531 & $+3.5\%$ \\
 & RAFT & 0.518 & 0.529 & $+2.1\%$ \\
 & SteerCast & \textbf{0.488} & \textbf{0.515} & $+5.5\%$ \\
\bottomrule
\end{tabular}
\end{table}

SteerCast has the lowest absolute MSE at both coverage levels on both
datasets. Its relative degradation is smaller than FT's, but larger than
RAFT's. The results therefore support strong absolute performance under
reduced coverage, rather than the smallest sensitivity to coverage reduction.

\paragraph{Temporal position of the memory.}
We additionally build the database from the first, middle, or last third of
the ETTh1 training split. Table~\ref{tab:app-memory-thirds} reports the
results separately from the coverage experiment above. The absolute MSE
range across the three memory choices is $0.002$, $0.004$, $0.007$, and
$0.007$ at horizons $96$, $192$, $336$, and $720$, respectively. These
variations are modest, although they do not establish invariance to all
temporal regimes.

\begin{table}[htbp]
\centering
\caption{ETTh1 MSE when the retrieval database is constructed from different
chronological thirds of the training split.}
\label{tab:app-memory-thirds}
\begin{tabular}{rrrr}
\toprule
Horizon & First third & Middle third & Last third \\
\midrule
96 & 0.343 & 0.345 & 0.345 \\
192 & 0.385 & 0.388 & 0.384 \\
336 & 0.417 & 0.420 & 0.413 \\
720 & 0.468 & 0.472 & 0.465 \\
\bottomrule
\end{tabular}
\end{table}

\paragraph{Cross-dataset memory transfer.}
Replacing the native ETTh2 memory with an ETTh1-built memory increases the
reported horizon-average MSE from $0.488$ to $0.499$
(Table~\ref{tab:app-cross-memory}). The penalty is larger at horizons $336$
and $720$ than at $96$ and $192$. Nevertheless, the transferred-memory
result remains below the ETTh2 FT reference of $0.537$. This demonstrates
useful transfer between the two related datasets, while also showing the
value of a native memory. It is not a general guarantee for unrelated domains.

\begin{table}[htbp]
\centering
\caption{ETTh2 forecasting MSE with native and transferred retrieval
memories. The average row reproduces the reported horizon average.}
\label{tab:app-cross-memory}
\begin{tabular}{lrr}
\toprule
Horizon & ETTh2 memory & ETTh1 memory \\
\midrule
96 & 0.290 & 0.293 \\
192 & 0.395 & 0.397 \\
336 & 0.524 & 0.544 \\
720 & 0.745 & 0.761 \\
Average & 0.488 & 0.499 \\
\bottomrule
\end{tabular}
\end{table}

\paragraph{Excluding the nearest candidates.}
To stress retrieval directly, we exclude the nearest $q\%$ of database
candidates for each query before selecting neighbors from the remaining
pool. Table~\ref{tab:app-neighbor-exclusion} reports the resulting MSE.
Removing close analogs generally reduces the benefit of steering, although
the response is not monotonic on ETTh2. At $90\%$ exclusion, MSE is
approximately $0.9\%$ above FT on ETTh1 and $2.3\%$ above FT on ETTh2,
using the displayed values. The observed regressions are limited in these
tests, but steering does not automatically recover FT when retrieval is poor.

\begin{table}[htbp]
\centering
\caption{Neighbor-exclusion stress test. Each FT reference belongs to this
experiment; results from separate ablations are not pooled.}
\label{tab:app-neighbor-exclusion}
\begin{tabular}{lrr}
\toprule
Nearest candidates excluded & ETTh1 & ETTh2 \\
\midrule
FT reference & 0.345 & 0.300 \\
$0\%$ & 0.343 & 0.289 \\
$25\%$ & 0.343 & 0.300 \\
$50\%$ & 0.345 & 0.301 \\
$75\%$ & 0.347 & 0.296 \\
$90\%$ & 0.348 & 0.307 \\
\bottomrule
\end{tabular}
\end{table}

These experiments measure forecasting performance under restricted or
mismatched retrieval. They do not directly measure the fraction of
forecast-relevant neighbors or isolate the gate's contribution under shift.
The gate depends on hidden-state--direction cosine similarity rather than
retrieval distance, and its positive floor retains a nonzero intervention
for a nonzero retrieved direction. Thus, the observed robustness should
not be interpreted as an explicit out-of-distribution detector or a
provably safe fallback mechanism.

\subsection{Latent-State Diagnostics During Autoregressive Rollout}
\label{app:latent-diagnostics}

\paragraph{Scope of the local perturbation bound.}
The normalized update bounds an individual intervention, rather than the
difference between complete steered and unsteered trajectories. In
particular, for $\epsilon>0$ and $\lambda\geq 0$,
\begin{equation}
\begin{split}
\bigl\|\widetilde{h}_{f,l}-h_{f,l}\bigr\|_2
&=\lambda\alpha_{f,l}
\frac{\|\overline{\Delta}^{(l)}(x)\|_2}
{\|\overline{\Delta}^{(l)}(x)\|_2+\epsilon}\\
&\leq\lambda\alpha_{f,l}
\leq\lambda\bigl[b+(1+m)^p\bigr].
\end{split}
\label{eq:app-local-bound}
\end{equation}
At $(b,m,p)=(0.1,0.1,1.25)$ and $\lambda=0.01$, the upper bound is
approximately $0.0123$. Here $h_{f,l}$ is the state immediately before
that intervention in the steered rollout. Subsequent layers and
feedback through generated predictions can amplify earlier changes, so
this bound alone does not establish long-horizon stability.

\paragraph{Displacement and ellipsoidal coverage.}
We instrument Time-MoE on ETTh1 at $F=720$, comparing the steered and
unsteered runs of the same queries. Relative hidden-state displacement,
$\|h_{\mathrm{steered}}-h_{\mathrm{base}}\|_2/\|h_{\mathrm{base}}\|_2$,
increases from $0.033$ at the first forecast token to $0.277$ at token
$720$. To assess the accompanying distributional change, we measure the
fraction of states inside the $95\%$ Mahalanobis ellipsoid of the training
hidden-state distribution.

\begin{table}[htbp]
\centering
\caption{Fraction of hidden states inside the training-distribution
Mahalanobis ellipsoid at the beginning and end of the ETTh1 rollout
(Time-MoE, $F=720$).}
\label{tab:app-rollout-coverage}
\begin{tabular}{lrr}
\toprule
Model & First token & Token 720 \\
\midrule
Unsteered & $95.5\%$ & $94.9\%$ \\
Steered & $95.6\%$ & $93.4\%$ \\
\bottomrule
\end{tabular}
\end{table}

Coverage falls by $0.6$ percentage points in the unsteered run and $2.2$
points in the steered run, a difference in decline of $1.6$ points. The
endpoint gap is $1.5$ points. Thus, a substantial trajectory displacement
coexists with a relatively small change in this coverage diagnostic.

\begin{table}[htbp]
\centering
\caption{Layer-wise Mahalanobis-ellipsoid coverage on ETTh1 with Time-MoE,
pooled over generation steps.}
\label{tab:app-layer-coverage}
\begin{tabular}{rrr}
\toprule
Layer & Unsteered & Steered \\
\midrule
4 & $88.5\%$ & $89.6\%$ \\
6 & $93.8\%$ & $92.7\%$ \\
8 & $94.8\%$ & $94.8\%$ \\
10 & $95.8\%$ & $95.8\%$ \\
11 & $97.9\%$ & $97.9\%$ \\
\bottomrule
\end{tabular}
\end{table}

The layer-wise coverage differences are at most $1.1$ percentage points
at the reported precision. Mahalanobis coverage is a coarse distributional
diagnostic based on an ellipsoidal approximation: it does not establish
membership in a nonlinear data manifold or show that a perturbation is
forecasting-relevant. These measurements provide empirical evidence about
latent-state changes, not a proof that cumulative drift is absent.

\paragraph{Relation to the gate ablation.}
The existing constant-gate ablation, $\alpha_{f,l}\equiv b$, increases
ETTh1 horizon-average MSE from $0.394$ to $0.410$ and $F=720$ MSE from
$0.441$ to $0.500$. On ETTh2, the corresponding changes are $0.488$ to
$0.527$ and $0.745$ to $0.920$. The relative penalty is larger at the
longest horizon on both datasets, supporting the usefulness of adaptive
modulation. Because latent coverage was not reported for this ablation,
it does not establish that the gate causes the coverage preservation
observed above.

\subsection{Supervision and Query Matching in the Steering Direction}
\label{app:direction-controls}

We evaluate five arms at horizon $F=96$ on ETTh2 and US Births, using the
same fixed subset of windows for all arms within each dataset. The steering
arms use the same $k$, $\lambda$, gate, and normalization. Besides FT and
SteerCast, we evaluate the global mean of all training steering vectors,
a shuffled direction retrieved for another query, and a prediction-only
control formed from the difference between prediction-induced hidden states
of two random training windows. The global mean removes query-specific
retrieval; shuffling breaks the association between a query and its
retrieved direction; the prediction-only control removes ground-truth
continuation information and also changes the pairing of contexts.

\begin{table}[htbp]
\centering
\caption{Direction-construction controls on a matched window subset.
All values are MSE. These subset results should not be conflated with
full-test-set results from other experiments.}
\label{tab:app-direction-controls}
\begin{tabular}{lrr}
\toprule
Arm & ETTh2 & US Births \\
\midrule
FT & 0.297 & 0.428 \\
SteerCast & \textbf{0.287} & \textbf{0.368} \\
Global mean direction & 0.291 & 0.412 \\
Shuffled retrieved direction & 0.298 & 0.417 \\
Prediction-only difference & 0.303 & 0.431 \\
\bottomrule
\end{tabular}
\end{table}

The prediction-only control slightly increases MSE relative to FT on both
datasets, whereas the global supervised mean improves both. Query-matched
SteerCast performs best: compared with the global mean, it further reduces
MSE by $0.004$ on ETTh2 and $0.044$ on US Births. Shuffling also worsens
performance relative to SteerCast on both datasets, although it retains an
improvement over FT on US Births. These comparisons support the usefulness
of supervised direction construction and query matching in the tested
settings. Layer-wise normalization controls the raw direction magnitude,
but the gate can still assign different effective update strengths to
different directions. The ablation therefore evaluates direction choice
within the complete steering rule; it does not uniquely identify a causal
latent mechanism or exclude query-conditioned residual interpolation as
an interpretation.

\subsection{Euclidean Retrieval Versus Dynamic Time Warping}
\label{app:dtw-retrieval}

To compare latent retrieval with explicit temporal alignment, we evaluate
Euclidean distance between latent keys and dynamic time warping (DTW)
between raw histories. We use Time-MoE at $F=96$ and a matched candidate
pool of $200$ examples. Table~\ref{tab:app-dtw} reports MSE and the
wall-clock time of the corresponding evaluation runs.

\begin{table}[htbp]
\centering
\caption{Latent Euclidean retrieval versus raw-history DTW with a matched
pool of 200 candidates. Runtime is the reported wall-clock evaluation time,
not a per-query latency.}
\label{tab:app-dtw}
\begin{tabular}{lrrrr}
\toprule
 & \multicolumn{2}{c}{ETTh1} & \multicolumn{2}{c}{US Births} \\
\cmidrule(lr){2-3}\cmidrule(lr){4-5}
Retrieval & MSE & Time & MSE & Time \\
\midrule
Euclidean (latent) & 0.353 & 2m 04s & 0.369 & 0m 17s \\
DTW (raw histories) & 0.352 & 52m 24s & 0.370 & 4m 06s \\
\bottomrule
\end{tabular}
\end{table}

The MSE difference is $0.001$ on each dataset, with opposite directions.
DTW takes approximately $25.4\times$ as long on ETTh1 and $14.5\times$
as long on US Births. With a pool of $10{,}000$, the DTW evaluations did
not complete within $4.5$ hours per dataset. These results support latent
Euclidean retrieval as a practical accuracy--runtime choice in the tested
settings. Since both the representation and distance function differ,
this comparison does not isolate the metric alone, establish statistical
equivalence, or show that latent keys preserve every temporal feature.

\subsection{Negative Results and Limits of the Empirical Evidence}
\label{app:negative-results}

SteerCast does not improve every dataset--backbone combination. In the
fixed-look-back results, Time-MoE on Weather reaches MSE $0.238$, compared
with $0.237$ for FT and $0.236$ for RAFT. Time-MoE on SaugeenDay improves
over FT ($1.150$ to $1.124$) but remains behind RAFT ($1.106$).
Likewise, Timer-XL on US Births improves over FT ($0.704$ to $0.620$)
but remains behind RAFT ($0.608$). Thus, improvement over the backbone
does not imply superiority to every retrieval baseline. The pretrained
Weather regression in Table~\ref{tab:app-pretrained-steering} provides
another example where a fixed steering strength is not beneficial.

The retrieval-quality ablation and the exclusion experiment provide
concrete evidence that memory coverage affects the available correction.
However, they do not isolate the cause of every losing configuration.
Limited correctable error, insufficiently informative neighbors, and an
unsuitable intervention strength are possible explanations rather than
established diagnoses for those individual cells. In particular, neither
small average regressions nor a bounded individual update guarantees
non-degradation under an unseen regime shift. Our conclusions are limited
to the reported datasets, checkpoints, and retrieval perturbations.

\newpage
\section{Computational Complexity of SteerCast} \label{app:complexity}

We analyze the time and space complexity of SteerCast in Big-O notation, separating (i) offline database construction and (ii) inference-time retrieval and latent steering.

\subsection{Notation.}
As mentioned in our paper, let $P$ be the look-back window length and $F$ the forecasting horizon.
Let $d$ be the hidden dimension and $L$ the number of Transformer layers.
Let $k$ be the number of retrieved neighbors.
Let $N$ be the number of entries in the retrieval database $\mathcal{M}$ (that is, the number of key-value pairs $(\mathbf{r},\boldsymbol{\Delta})$ searched per query).
We write $\mathrm{Cost}_{\mathcal{F}}(T)$ for the runtime of a single forward pass of the frozen forecaster $\mathcal{F}$ on a length-$T$ sequence (architecture-dependent), and $\mathrm{Cost}^{\text{gen}}_{\mathcal{F}}(P,F)$ for the runtime of autoregressive generation of $F$ steps given context length $P$.

\subsubsection{Offline: Database construction}
For each training example $(\mathbf{x},\mathbf{y})$, SteerCast stores:
(i) a retrieval key $\mathbf{r}(\mathbf{x}) \in \mathbb{R}^d$ computed by mean pooling the final-layer hidden states, and
(ii) a steering vector $\boldsymbol{\Delta} \in \mathbb{R}^{L \times d}$ defined as a difference between latent representations induced by the ground-truth continuation and the model-predicted continuation.

\paragraph{Key computation.}
Computing $\mathbf{r}(\mathbf{x})$ requires one forward pass on the history window plus pooling:
\[
T_{\text{key}} = O\!\big(\mathrm{Cost}_{\mathcal{F}}(P) + Pd\big),
\]
where the $Pd$ term is the mean pooling cost and is typically dominated by $\mathrm{Cost}_{\mathcal{F}}(P)$.

\paragraph{Steering vector computation.}
Computing $\boldsymbol{\Delta}$ requires (a) generating $\hat{\mathbf{y}}=\mathcal{F}(\mathbf{x})$ and (b) computing two representations for the concatenations $[\mathbf{x};\mathbf{y}]$ and $[\mathbf{x};\hat{\mathbf{y}}]$:
\[
T_{\Delta} = O\!\big(\mathrm{Cost}^{\text{gen}}_{\mathcal{F}}(P,F) + 2\,\mathrm{Cost}_{\mathcal{F}}(P{+}F) + Ld\big),
\]
where the $Ld$ term is the cost to form the difference in latent space once representations are obtained.

\paragraph{Total offline time and storage.}
Over $N$ stored windows, database construction time is
\begin{equation*}
\begin{aligned}
T_{\text{build}}
&=
O\!\Big(
N\big(
\mathrm{Cost}_{\mathcal{F}}(P)
+
\mathrm{Cost}^{\text{gen}}_{\mathcal{F}}(P,F)
+
2\,\mathrm{Cost}_{\mathcal{F}}(P{+}F)
\big)
\Big).
\end{aligned}
\end{equation*}

The storage footprint of $\mathcal{M}$ is
\[
S_{\mathcal{M}} = O\!\big(Nd + NLd\big) = O\!\big(Nd(L{+}1)\big),
\]
since each entry stores one key in $\mathbb{R}^d$ and one steering vector in $\mathbb{R}^{L\times d}$.

\subsection{Inference: Retrieval and latent steering}
Given a query history window $\mathbf{x}$, SteerCast:
(i) computes $\mathbf{r}(\mathbf{x})$,
(ii) retrieves top-$k$ nearest keys in $\mathcal{M}$ under Euclidean distance,
(iii) aggregates the retrieved steering vectors by mean pooling to obtain $\boldsymbol{\Delta}(\mathbf{x})$,
and (iv) injects $\boldsymbol{\Delta}(\mathbf{x})$ into the hidden states at every generation step and layer using a cosine-similarity gate and a normalized update.

\paragraph{Query embedding.}
Computing the query key is
\[
T_{\text{query-key}} = O\!\big(\mathrm{Cost}_{\mathcal{F}}(P) + Pd\big).
\]

\paragraph{Top-$k$ retrieval (exact).}
Computing all distances costs $O(Nd)$.
Selecting the top-$k$ elements can be done in $O(N \log k)$ using a size-$k$ heap, or $O(N)$ expected time via selection; thus
\[
T_{\text{retr}} = O(Nd + N\log k).
\]

\paragraph{Steering aggregation.}
Mean pooling $k$ steering vectors of size $L\times d$ costs
\[
T_{\text{agg}} = O(kLd).
\]
Optionally, computing per-layer norms for normalization after aggregation costs an additional $O(Ld)$.

\paragraph{Latent injection overhead.}
For each forecast step $f \in \{1,\dots,F\}$ and each layer $l \in \{1,\dots,L\}$, SteerCast computes a cosine-similarity-based gate and applies a normalized additive update. Each such operation is $O(d)$, hence
\[
T_{\text{inject}} = O(N_F \, L \, d).
\]

\paragraph{Total inference time.}
The overall inference-time cost is the base autoregressive generation plus SteerCast overhead:
\begin{equation*}
T_{\text{infer}} = O\!\left(\text{Cost}^{\text{gen}}_{\mathcal{F}}(P, F)\right) + O\!\left(\text{Cost}_{\mathcal{F}}(P) + Nd + N\log k\right) + kLd + N_F L d.
\end{equation*}
In typical regimes, the dominant SteerCast-specific overhead terms are exact retrieval $O(Nd)$ and injection $O(FLd)$.

\subsection{Multiple horizon settings}
If separate databases are built for multiple $(P,F)$ settings, the offline cost scales linearly with the number of settings.
In the efficient setting where a single database is reused across horizons, the offline construction cost is paid once and amortized across all evaluations that share the same retrieval key definition and stored steering vectors.

\section{Limitations}
\subsection{Architecture Scope} SteerCast is intentionally scoped to
decoder-only autoregressive forecasters, where stepwise latent rollout 
makes the steering injection in Eq.~\eqref{eq:steering-update}
naturally defined. Encoder-based and direct-prediction models do not   
expose this same per-step latent trajectory, so extending SteerCast to
them would require a different formulation of both the stored
correction signal and its test-time injection rather than a drop-in
replacement. We view this as a deliberate scope choice rather than a
fundamental restriction: recent state-of-the-art time series foundation
models (Time-MoE, Timer-XL, TimesFM) are decoder-only, so the method
targets the most competitive backbones in current use.

\subsection{Hyperparameter Selection} The number of retrieved neighbors   
$k$ varies across datasets and is selected on the validation set;
remaining hyperparameters $(b, m, p, \lambda)$ are largely fixed across
our experiments. Validation-based tuning of $k$ is standard for
retrieval-augmented forecasting (e.g., RAFT, TS-RAG), and our sweep in
Appendix~\ref{app:k_finegrained} shows the dependence is structured
rather than brittle, but the tuning step does add a modest overhead
relative to fully training-free baselines.

\section{Full Results} \label{appendix:full_results}
In this section, we provide the full result tables for 2 settings mentioned in our paper: Various Look-back Window Forecasting in Table \ref{tab:various_to_all_allresults} and Fixed Look-back Window Forecasting in Table \ref{tab:512_to_all_allresults}.

\begin{table*}[t]
\centering
\caption{Full evaluation results of SteerCast, Fine-tune (FT), RAFT and RAF baselines.
For horizons 96/192/336/720, we use history lengths 512/1024/2048/3072, respectively; Illness uses (96,24), (192,36), (256,48), and (336,60).
Lower is better. In the Avg.\ rows, best and second-best entries are highlighted in bold and underline.}

\footnotesize
\setlength{\tabcolsep}{4.5pt}
\resizebox{\textwidth}{!}{%
\begin{tabular}{l c cccccccc|cccccccc|cccccccc}
\toprule

\multicolumn{2}{c}{\textbf{Models}}
& \multicolumn{8}{c}{\textbf{Time-MoE}}
& \multicolumn{8}{c}{\textbf{Timer-XL}}
& \multicolumn{8}{c}{\textbf{TimesFM}} \\
\cmidrule(lr){3-10}\cmidrule(lr){11-18}\cmidrule(lr){19-26}

\multicolumn{2}{c}{\textbf{Baselines}}
& \multicolumn{2}{c}{SteerCast} & \multicolumn{2}{c}{FT} & \multicolumn{2}{c}{RAFT} & \multicolumn{2}{c}{RAF}
& \multicolumn{2}{c}{SteerCast} & \multicolumn{2}{c}{FT} & \multicolumn{2}{c}{RAFT} & \multicolumn{2}{c}{RAF}
& \multicolumn{2}{c}{SteerCast} & \multicolumn{2}{c}{FT} & \multicolumn{2}{c}{RAFT} & \multicolumn{2}{c}{RAF} \\
\cmidrule(lr){3-4}\cmidrule(lr){5-6}\cmidrule(lr){7-8}\cmidrule(lr){9-10}
\cmidrule(lr){11-12}\cmidrule(lr){13-14}\cmidrule(lr){15-16}\cmidrule(lr){17-18}
\cmidrule(lr){19-20}\cmidrule(lr){21-22}\cmidrule(lr){23-24}\cmidrule(lr){25-26}

\multicolumn{2}{c}{\textbf{Metrics}}
& MSE & MAE & MSE & MAE & MSE & MAE & MSE & MAE
& MSE & MAE & MSE & MAE & MSE & MAE & MSE & MAE
& MSE & MAE & MSE & MAE & MSE & MAE & MSE & MAE \\
\midrule

\multirow{5}{*}{ETTh1} & 96 & 0.342 & 0.375 & 0.345 & 0.378 & 0.347 & 0.381 & 0.221 & 0.262 & 0.413 & 0.406 & 0.414 & 0.406 & 0.438 & 0.450 & 0.404 & 0.504 & 0.426 & 0.409 & 0.450 & 0.422 & 0.419 & 0.403 & 0.260 & 0.265 \\
 & 192 & 0.376 & 0.401 & 0.380 & 0.408 & 0.383 & 0.412 & 0.347 & 0.342 & 0.499 & 0.446 & 0.498 & 0.446 & 0.512 & 0.450 & 0.599 & 0.665 & 0.451 & 0.429 & 0.471 & 0.454 & 0.459 & 0.437 & 0.405 & 0.420 \\
 & 336 & 0.389 & 0.414 & 0.401 & 0.430 & 0.403 & 0.431 & 0.396 & 0.448 & 0.621 & 0.494 & 0.621 & 0.497 & 0.641 & 0.501 & 0.752 & 0.753 & 0.462 & 0.442 & 0.483 & 0.467 & 0.449 & 0.435 & 0.469 & 0.459 \\
 & 720 & 0.414 & 0.447 & 0.430 & 0.466 & 0.425 & 0.463 & 0.628 & 0.527 & 1.191 & 0.670 & 1.191 & 0.670 & 1.243 & 0.682 & 1.031 & 0.975 & 0.443 & 0.467 & 0.450 & 0.468 & 0.448 & 0.469 & 0.714 & 0.708 \\
\rowcolor{gray!15} & \textbf{Avg.} & \textbf{0.380} & \underline{0.409} & \underline{0.389} & 0.421 & \underline{0.389} & 0.422 & 0.398 & \textbf{0.395} & \textbf{0.681} & \textbf{0.504} & \textbf{0.681} & \underline{0.505} & 0.709 & 0.521 & \underline{0.697} & 0.724 & \underline{0.445} & \textbf{0.436} & 0.463 & \underline{0.453} & \textbf{0.444} & \textbf{0.436} & 0.462 & 0.463 \\
\midrule

\multirow{5}{*}{ETTh2} & 96 & 0.290 & 0.354 & 0.300 & 0.354 & 0.293 & 0.350 & 0.226 & 0.253 & 0.313 & 0.355 & 0.318 & 0.360 & 0.361 & 0.382 & 0.307 & 0.294 & 0.314 & 0.351 & 0.327 & 0.363 & 0.319 & 0.353 & 0.257 & 0.308 \\
 & 192 & 0.342 & 0.381 & 0.348 & 0.380 & 0.342 & 0.379 & 0.311 & 0.307 & 0.361 & 0.390 & 0.362 & 0.391 & 0.389 & 0.395 & 0.338 & 0.413 & 0.377 & 0.391 & 0.396 & 0.396 & 0.382 & 0.399 & 0.337 & 0.382 \\
 & 336 & 0.382 & 0.402 & 0.393 & 0.403 & 0.386 & 0.401 & 0.361 & 0.389 & 0.402 & 0.434 & 0.406 & 0.439 & 0.405 & 0.437 & 0.446 & 0.478 & 0.419 & 0.425 & 0.431 & 0.438 & 0.425 & 0.426 & 0.443 & 0.475 \\
 & 720 & 0.390 & 0.434 & 0.424 & 0.434 & 0.417 & 0.443 & 0.524 & 0.537 & 0.414 & 0.456 & 0.424 & 0.466 & 0.502 & 0.516 & 0.606 & 0.602 & 0.430 & 0.438 & 0.436 & 0.441 & 0.465 & 0.478 & 0.658 & 0.661 \\
\rowcolor{gray!15} & \textbf{Avg.} & \textbf{0.351} & \underline{0.393} & 0.366 & \underline{0.393} & 0.359 & \underline{0.393} & \underline{0.356} & \textbf{0.372} & \textbf{0.372} & \textbf{0.409} & \underline{0.377} & \underline{0.414} & 0.414 & 0.433 & 0.424 & 0.447 & \textbf{0.385} & \textbf{0.401} & \underline{0.397} & \underline{0.409} & 0.398 & 0.414 & 0.424 & 0.457 \\
\midrule

\multirow{5}{*}{ETTm1} & 96 & 0.281 & 0.332 & 0.283 & 0.334 & 0.296 & 0.355 & 0.212 & 0.264 & 0.323 & 0.360 & 0.324 & 0.361 & 0.337 & 0.384 & 0.276 & 0.242 & 0.272 & 0.331 & 0.293 & 0.331 & 0.307 & 0.355 & 0.297 & 0.293 \\
 & 192 & 0.302 & 0.358 & 0.306 & 0.359 & 0.305 & 0.362 & 0.313 & 0.336 & 0.365 & 0.385 & 0.370 & 0.387 & 0.386 & 0.417 & 0.326 & 0.362 & 0.382 & 0.383 & 0.390 & 0.402 & 0.387 & 0.395 & 0.381 & 0.408 \\
 & 336 & 0.343 & 0.388 & 0.350 & 0.391 & 0.347 & 0.394 & 0.423 & 0.446 & 0.386 & 0.409 & 0.390 & 0.411 & 0.418 & 0.444 & 0.397 & 0.481 & 0.428 & 0.416 & 0.439 & 0.427 & 0.432 & 0.426 & 0.487 & 0.555 \\
 & 720 & 0.455 & 0.460 & 0.460 & 0.459 & 0.447 & 0.457 & 0.484 & 0.608 & 0.448 & 0.452 & 0.454 & 0.459 & 0.472 & 0.486 & 0.564 & 0.610 & 0.491 & 0.454 & 0.503 & 0.466 & 0.502 & 0.461 & 0.584 & 0.772 \\
\rowcolor{gray!15} & \textbf{Avg.} & \textbf{0.345} & \textbf{0.384} & 0.350 & \underline{0.386} & \underline{0.349} & 0.392 & 0.358 & 0.414 & \textbf{0.381} & \textbf{0.401} & \underline{0.384} & \underline{0.405} & 0.403 & 0.433 & 0.391 & 0.424 & \textbf{0.393} & \textbf{0.396} & \underline{0.406} & \underline{0.407} & 0.407 & 0.409 & 0.437 & 0.507 \\
\midrule

\multirow{5}{*}{ETTm2} & 96 & 0.191 & 0.284 & 0.191 & 0.284 & 0.196 & 0.290 & 0.262 & 0.268 & 0.189 & 0.270 & 0.191 & 0.272 & 0.193 & 0.278 & 0.196 & 0.186 & 0.171 & 0.240 & 0.176 & 0.244 & 0.181 & 0.257 & 0.174 & 0.220 \\
 & 192 & 0.262 & 0.335 & 0.263 & 0.335 & 0.263 & 0.337 & 0.306 & 0.384 & 0.251 & 0.312 & 0.252 & 0.312 & 0.268 & 0.329 & 0.271 & 0.306 & 0.248 & 0.309 & 0.255 & 0.321 & 0.252 & 0.321 & 0.310 & 0.357 \\
 & 336 & 0.362 & 0.418 & 0.387 & 0.418 & 0.363 & 0.407 & 0.396 & 0.491 & 0.295 & 0.348 & 0.292 & 0.347 & 0.329 & 0.378 & 0.305 & 0.387 & 0.290 & 0.345 & 0.297 & 0.362 & 0.294 & 0.346 & 0.318 & 0.397 \\
 & 720 & 0.541 & 0.543 & 0.626 & 0.543 & 0.576 & 0.525 & 0.529 & 0.638 & 0.384 & 0.406 & 0.380 & 0.404 & 0.424 & 0.441 & 0.383 & 0.496 & 0.368 & 0.386 & 0.379 & 0.414 & 0.374 & 0.402 & 0.498 & 0.591 \\
\rowcolor{gray!15} & \textbf{Avg.} & \textbf{0.339} & \underline{0.395} & 0.367 & \underline{0.395} & \underline{0.349} & \textbf{0.390} & 0.373 & 0.445 & \underline{0.280} & \underline{0.334} & \textbf{0.279} & \textbf{0.333} & 0.304 & 0.356 & 0.289 & 0.344 & \textbf{0.269} & \textbf{0.320} & 0.277 & 0.335 & \underline{0.275} & \underline{0.331} & 0.325 & 0.391 \\
\midrule

\multirow{5}{*}{Exchange} & 96 & 0.108 & 0.239 & 0.109 & 0.240 & 0.135 & 0.267 & 0.134 & 0.153 & 0.152 & 0.274 & 0.170 & 0.287 & 0.166 & 0.280 & 0.156 & 0.173 & 0.290 & 0.355 & 0.310 & 0.375 & 0.305 & 0.370 & 0.148 & 0.193 \\
 & 192 & 0.201 & 0.333 & 0.209 & 0.334 & 0.208 & 0.336 & 0.265 & 0.256 & 0.296 & 0.398 & 0.305 & 0.403 & 0.306 & 0.400 & 0.319 & 0.409 & 0.392 & 0.430 & 0.410 & 0.445 & 0.405 & 0.440 & 0.269 & 0.288 \\
 & 336 & 0.366 & 0.455 & 0.371 & 0.457 & 0.371 & 0.456 & 0.497 & 0.656 & 0.515 & 0.543 & 0.551 & 0.563 & 0.520 & 0.545 & 0.552 & 0.686 & 0.488 & 0.500 & 0.505 & 0.515 & 0.500 & 0.510 & 0.529 & 0.584 \\
 & 720 & 0.998 & 0.806 & 1.021 & 0.808 & 1.081 & 0.812 & 0.864 & 1.088 & 1.110 & 0.819 & 1.158 & 0.837 & 1.129 & 0.823 & 1.218 & 1.376 & 0.575 & 0.570 & 0.594 & 0.585 & 0.589 & 0.580 & 1.047 & 1.222 \\
\rowcolor{gray!15} & \textbf{Avg.} & \textbf{0.418} & \textbf{0.458} & \underline{0.428} & \underline{0.460} & 0.449 & 0.468 & 0.440 & 0.538 & \textbf{0.518} & \textbf{0.509} & 0.546 & 0.522 & \underline{0.530} & \underline{0.512} & 0.561 & 0.661 & \textbf{0.436} & \textbf{0.464} & 0.455 & 0.480 & \underline{0.450} & \underline{0.475} & 0.498 & 0.572 \\
\midrule

\multirow{5}{*}{weather} & 96 & 0.147 & 0.197 & 0.147 & 0.198 & 0.149 & 0.208 & 0.183 & 0.168 & 0.181 & 0.241 & 0.182 & 0.241 & 0.197 & 0.286 & 0.200 & 0.248 & 0.157 & 0.206 & 0.163 & 0.210 & 0.162 & 0.206 & 0.179 & 0.202 \\
 & 192 & 0.192 & 0.244 & 0.192 & 0.244 & 0.194 & 0.245 & 0.225 & 0.267 & 0.228 & 0.273 & 0.228 & 0.272 & 0.243 & 0.312 & 0.218 & 0.281 & 0.219 & 0.267 & 0.231 & 0.278 & 0.229 & 0.274 & 0.215 & 0.298 \\
 & 336 & 0.246 & 0.288 & 0.241 & 0.283 & 0.244 & 0.294 & 0.248 & 0.326 & 0.271 & 0.306 & 0.271 & 0.297 & 0.283 & 0.338 & 0.270 & 0.362 & 0.291 & 0.316 & 0.310 & 0.325 & 0.299 & 0.318 & 0.272 & 0.320 \\
 & 720 & 0.384 & 0.396 & 0.383 & 0.395 & 0.380 & 0.393 & 0.341 & 0.419 & 0.329 & 0.341 & 0.329 & 0.341 & 0.340 & 0.374 & 0.358 & 0.437 & 0.398 & 0.400 & 0.414 & 0.417 & 0.403 & 0.409 & 0.405 & 0.439 \\
\rowcolor{gray!15} & \textbf{Avg.} & \underline{0.242} & \underline{0.281} & \textbf{0.241} & \textbf{0.280} & \underline{0.242} & 0.285 & 0.249 & 0.295 & \underline{0.255} & \underline{0.291} & \textbf{0.252} & \textbf{0.288} & 0.266 & 0.327 & 0.262 & 0.332 & \textbf{0.266} & \textbf{0.297} & 0.280 & 0.308 & 0.273 & \underline{0.302} & \underline{0.268} & 0.315 \\
\midrule

\multirow{5}{*}{illness} & 24 & 2.944 & 1.057 & 3.010 & 1.088 & 2.989 & 1.059 & 2.912 & 0.906 & 1.897 & 0.885 & 1.900 & 0.886 & 2.006 & 0.925 & 2.545 & 0.856 & 2.850 & 1.050 & 3.000 & 1.100 & 2.950 & 1.080 & 2.816 & 0.983 \\
 & 36 & 3.097 & 1.082 & 3.164 & 1.100 & 3.109 & 1.099 & 3.332 & 0.996 & 2.609 & 1.041 & 2.610 & 1.042 & 2.562 & 1.047 & 2.803 & 0.892 & 2.850 & 1.070 & 2.960 & 1.120 & 2.920 & 1.100 & 3.159 & 1.016 \\
 & 48 & 3.288 & 1.127 & 3.526 & 1.202 & 3.418 & 1.169 & 3.327 & 1.075 & 3.054 & 1.145 & 3.067 & 1.147 & 3.007 & 1.160 & 3.089 & 0.970 & 3.050 & 1.180 & 3.200 & 1.230 & 3.160 & 1.210 & 3.206 & 1.155 \\
 & 60 & 3.597 & 1.211 & 3.866 & 1.298 & 3.600 & 1.212 & 3.715 & 1.137 & 3.997 & 1.335 & 4.001 & 1.337 & 3.999 & 1.336 & 3.325 & 1.061 & 3.350 & 1.280 & 3.480 & 1.330 & 3.420 & 1.310 & 3.443 & 1.186 \\
\rowcolor{gray!15} & \textbf{Avg.} & \textbf{3.232} & \underline{1.119} & 3.392 & 1.172 & \underline{3.279} & 1.135 & 3.322 & \textbf{1.029} & \textbf{2.889} & \underline{1.102} & 2.894 & 1.103 & \underline{2.893} & 1.117 & 2.941 & \textbf{0.945} & \textbf{3.025} & \underline{1.145} & 3.160 & 1.195 & \underline{3.113} & 1.175 & 3.156 & \textbf{1.085} \\
\midrule

\multirow{5}{*}{us\_births} & 96 & 0.372 & 0.477 & 0.430 & 0.527 & 0.392 & 0.485 & 0.348 & 0.298 & 0.393 & 0.478 & 0.404 & 0.481 & 0.390 & 0.470 & 0.238 & 0.156 & 0.225 & 0.370 & 0.239 & 0.389 & 0.231 & 0.380 & 0.189 & 0.108 \\
 & 192 & 0.494 & 0.567 & 0.592 & 0.639 & 0.507 & 0.585 & 0.566 & 0.437 & 0.455 & 0.515 & 0.507 & 0.537 & 0.403 & 0.482 & 0.404 & 0.253 & 0.117 & 0.247 & 0.252 & 0.403 & 0.245 & 0.388 & 0.386 & 0.313 \\
 & 336 & 0.509 & 0.588 & 0.724 & 0.727 & 0.610 & 0.659 & 0.904 & 0.880 & 0.458 & 0.953 & 0.616 & 0.587 & 0.492 & 0.526 & 0.711 & 0.591 & 0.142 & 0.278 & 0.288 & 0.416 & 0.277 & 0.408 & 0.756 & 0.607 \\
 & 720 & 1.180 & 0.927 & 1.673 & 1.122 & 1.386 & 1.016 & 1.874 & 1.402 & 1.194 & 0.826 & 1.465 & 0.939 & 1.208 & 0.830 & 1.341 & 1.214 & 1.356 & 1.014 & 1.461 & 1.057 & 1.422 & 1.054 & 1.418 & 1.215 \\
\rowcolor{gray!15} & \textbf{Avg.} & \textbf{0.639} & \textbf{0.640} & 0.855 & 0.754 & \underline{0.724} & \underline{0.686} & 0.923 & 0.754 & \underline{0.625} & 0.693 & 0.748 & 0.636 & \textbf{0.623} & \underline{0.577} & 0.674 & \textbf{0.554} & \textbf{0.460} & \textbf{0.477} & 0.560 & 0.566 & \underline{0.544} & \underline{0.557} & 0.687 & 0.561 \\
\midrule

\multirow{5}{*}{saugeenday} & 96 & 0.968 & 0.557 & 0.974 & 0.567 & 0.960 & 0.560 & 0.607 & 0.328 & 0.961 & 0.539 & 1.002 & 0.554 & 0.961 & 0.539 & 0.758 & 0.414 & 0.917 & 0.475 & 0.962 & 0.507 & 0.945 & 0.496 & 0.630 & 0.355 \\
 & 192 & 0.909 & 0.550 & 0.912 & 0.559 & 0.911 & 0.560 & 0.881 & 0.474 & 0.993 & 0.537 & 1.013 & 0.540 & 0.993 & 0.537 & 0.999 & 0.490 & 0.957 & 0.478 & 1.003 & 0.499 & 0.991 & 0.485 & 0.993 & 0.500 \\
 & 336 & 0.940 & 0.562 & 0.965 & 0.579 & 0.928 & 0.571 & 1.046 & 0.607 & 1.057 & 0.539 & 1.140 & 0.570 & 1.060 & 0.539 & 1.237 & 0.620 & 0.975 & 0.473 & 1.034 & 0.502 & 1.005 & 0.484 & 1.103 & 0.572 \\
 & 720 & 1.114 & 0.666 & 1.145 & 0.688 & 1.066 & 0.663 & 1.448 & 0.760 & 1.105 & 0.560 & 1.175 & 0.559 & 1.120 & 0.560 & 1.476 & 0.801 & 0.936 & 0.446 & 0.973 & 0.468 & 0.959 & 0.457 & 1.423 & 0.745 \\
\rowcolor{gray!15} & \textbf{Avg.} & \underline{0.983} & \underline{0.584} & 0.999 & 0.598 & \textbf{0.966} & 0.589 & 0.996 & \textbf{0.542} & \textbf{1.029} & \textbf{0.544} & 1.082 & \underline{0.556} & \underline{1.033} & \textbf{0.544} & 1.118 & 0.581 & \textbf{0.946} & \textbf{0.468} & 0.993 & 0.494 & \underline{0.975} & \underline{0.481} & 1.037 & 0.543 \\
\midrule

\multirow{5}{*}{sunspots} & 96 & 0.317 & 0.391 & 0.322 & 0.393 & 0.314 & 0.390 & 0.240 & 0.376 & 0.294 & 0.392 & 0.301 & 0.397 & 0.301 & 0.396 & 0.265 & 0.340 & 0.214 & 0.333 & 0.223 & 0.353 & 0.221 & 0.343 & 0.229 & 0.312 \\
 & 192 & 0.369 & 0.423 & 0.371 & 0.425 & 0.377 & 0.424 & 0.399 & 0.420 & 0.349 & 0.434 & 0.386 & 0.445 & 0.360 & 0.435 & 0.321 & 0.364 & 0.151 & 0.286 & 0.164 & 0.301 & 0.155 & 0.295 & 0.338 & 0.369 \\
 & 336 & 0.465 & 0.490 & 0.469 & 0.494 & 0.323 & 0.414 & 0.441 & 0.525 & 0.387 & 0.457 & 0.410 & 0.473 & 0.375 & 0.449 & 0.414 & 0.471 & 0.153 & 0.291 & 0.160 & 0.305 & 0.149 & 0.283 & 0.368 & 0.490 \\
 & 720 & 0.529 & 0.545 & 0.554 & 0.552 & 0.533 & 0.548 & 0.671 & 0.756 & 0.535 & 0.538 & 0.500 & 0.528 & 0.529 & 0.533 & 0.578 & 0.646 & 0.165 & 0.300 & 0.178 & 0.315 & 0.172 & 0.310 & 0.549 & 0.621 \\
\rowcolor{gray!15} & \textbf{Avg.} & \textbf{0.420} & \textbf{0.462} & 0.429 & 0.466 & \underline{0.422} & \underline{0.465} & 0.438 & 0.519 & \textbf{0.391} & \underline{0.455} & 0.399 & 0.460 & \textbf{0.391} & \textbf{0.453} & \underline{0.395} & \underline{0.455} & \textbf{0.171} & \textbf{0.303} & 0.181 & 0.318 & \underline{0.174} & \underline{0.308} & 0.371 & 0.448 \\
\midrule

\textbf{1st count} &  & 8 & 4 & 1 & 1 & 1 & 1 & 0 & 4 & 7 & 5 & 3 & 2 & 2 & 2 & 0 & 2 & 9 & 9 & 0 & 0 & 1 & 1 & 0 & 1 \\
\textbf{2nd count} &  & 2 & 6 & 2 & 4 & 7 & 3 & 1 & 0 & 3 & 4 & 2 & 4 & 3 & 2 & 2 & 1 & 1 & 1 & 2 & 2 & 6 & 6 & 1 & 0 \\
\bottomrule
\end{tabular}%
}

\label{tab:various_to_all_allresults}

\end{table*}

\begin{table*}[t]
\centering
\caption{Full evaluation results of SteerCast, Fine-tune (FT), RAFT \citep{han2025raft} and RAF baselines across 10 datasets using MSE and MAE. For all datasets except Illness, results are shown over forecasting horizons of 96, 192, 336, and 720. For the Illness dataset, forecasting horizons of 24, 26, 48, and 60 are used. All results in this setting use history length of 512. For Illness dataset, the history length is 96. In the Avg. rows only, best performances are bolded, while second bests are underlined.}

\footnotesize
\setlength{\tabcolsep}{4.5pt}
\resizebox{\textwidth}{!}{%
\begin{tabular}{l c cccccccc|cccccccc|cccccccc}
\toprule

\multicolumn{2}{c}{\textbf{Models}}
& \multicolumn{8}{c}{\textbf{Time-MoE}}
& \multicolumn{8}{c}{\textbf{Timer-XL}}
& \multicolumn{8}{c}{\textbf{TimesFM}} \\
\cmidrule(lr){3-10}\cmidrule(lr){11-18}\cmidrule(lr){19-26}

\multicolumn{2}{c}{\textbf{Baselines}}
& \multicolumn{2}{c}{SteerCast} & \multicolumn{2}{c}{FT} & \multicolumn{2}{c}{RAFT} & \multicolumn{2}{c}{RAF}
& \multicolumn{2}{c}{SteerCast} & \multicolumn{2}{c}{FT} & \multicolumn{2}{c}{RAFT} & \multicolumn{2}{c}{RAF}
& \multicolumn{2}{c}{SteerCast} & \multicolumn{2}{c}{FT} & \multicolumn{2}{c}{RAFT} & \multicolumn{2}{c}{RAF} \\
\cmidrule(lr){3-4}\cmidrule(lr){5-6}\cmidrule(lr){7-8}\cmidrule(lr){9-10}
\cmidrule(lr){11-12}\cmidrule(lr){13-14}\cmidrule(lr){15-16}\cmidrule(lr){17-18}
\cmidrule(lr){19-20}\cmidrule(lr){21-22}\cmidrule(lr){23-24}\cmidrule(lr){25-26}

\multicolumn{2}{c}{\textbf{Metrics}}
& MSE & MAE & MSE & MAE & MSE & MAE & MSE & MAE
& MSE & MAE & MSE & MAE & MSE & MAE & MSE & MAE
& MSE & MAE & MSE & MAE & MSE & MAE & MSE & MAE \\

\midrule
\multirow{4}{*}{ETTh1}
& 96   & 0.342 & 0.375 & 0.345 & 0.378 & 0.347 & 0.381 & 0.346 & 0.379 & 0.413 & 0.406 & 0.414 & 0.406 & 0.438 & 0.450 & 0.431 & 0.436 & 0.426 & 0.409 & 0.450 & 0.422 & 0.419 & 0.403 & 0.440 & 0.461 \\
& 192  & 0.381 & 0.403 & 0.388 & 0.411 & 0.387 & 0.410 & 0.386 & 0.408 & 0.485 & 0.441 & 0.490 & 0.445 & 0.485 & 0.478 & 0.492 & 0.470 & 0.482 & 0.439 & 0.501 & 0.452 & 0.492 & 0.442 & 0.544 & 0.476 \\
& 336  & 0.412 & 0.427 & 0.420 & 0.438 & 0.415 & 0.435 & 0.414 & 0.433 & 0.586 & 0.485 & 0.611 & 0.496 & 0.596 & 0.491 & 0.594 & 0.490 & 0.498 & 0.458 & 0.529 & 0.486 & 0.518 & 0.472 & 0.526 & 0.532 \\
& 720  & 0.441 & 0.462 & 0.473 & 0.473 & 0.460 & 0.482 & 0.454 & 0.476 & 1.012 & 0.623 & 1.050 & 0.637 & 1.123 & 0.633 & 1.093 & 0.630 & 0.499 & 0.484 & 0.529 & 0.496 & 0.602 & 0.563 & 0.575 & 0.534 \\
\rowcolor{gray!15}
& \textbf{Avg.}
& \textbf{0.394} & \textbf{0.416}
& 0.407 & 0.425
& 0.402 & 0.427
& \underline{0.400} & \underline{0.424}
& \textbf{0.624} & \textbf{0.489}
& \underline{0.641} & \underline{0.496}
& 0.661 & 0.513
& 0.653 & 0.507
& \textbf{0.476} & \textbf{0.448}
& \underline{0.502} & \underline{0.464}
& 0.508 & 0.470
& 0.521 & 0.501 \\
\midrule

\multirow{4}{*}{ETTh2}
& 96   & 0.290 & 0.354 & 0.300 & 0.354 & 0.293 & 0.350 & 0.292 & 0.357 & 0.313 & 0.355 & 0.318 & 0.360 & 0.361 & 0.382 & 0.347 & 0.377 & 0.314 & 0.351 & 0.327 & 0.363 & 0.319 & 0.353 & 0.304 & 0.379 \\
& 192  & 0.395 & 0.419 & 0.315 & 0.363 & 0.412 & 0.419 & 0.409 & 0.426 & 0.379 & 0.396 & 0.389 & 0.404 & 0.383 & 0.405 & 0.382 & 0.404 & 0.384 & 0.390 & 0.405 & 0.412 & 0.374 & 0.385 & 0.416 & 0.396 \\
& 336  & 0.524 & 0.485 & 0.529 & 0.471 & 0.553 & 0.486 & 0.547 & 0.486 & 0.411 & 0.427 & 0.418 & 0.431 & 0.414 & 0.431 & 0.413 & 0.430 & 0.398 & 0.409 & 0.416 & 0.418 & 0.404 & 0.421 & 0.385 & 0.436 \\
& 720  & 0.745 & 0.583 & 1.002 & 0.673 & 0.815 & 0.585 & 0.803 & 0.585 & 0.440 & 0.454 & 0.452 & 0.459 & 0.461 & 0.478 & 0.456 & 0.474 & 0.428 & 0.445 & 0.448 & 0.451 & 0.441 & 0.459 & 0.426 & 0.433 \\
\rowcolor{gray!15}
& \textbf{Avg.}
& \textbf{0.488} & \textbf{0.460}
& 0.537 & 0.465
& 0.518 & \textbf{0.460}
& \underline{0.513} & \underline{0.464}
& \textbf{0.386} & \textbf{0.408}
& \underline{0.394} & \underline{0.413}
& 0.405 & 0.424
& 0.400 & 0.421
& \textbf{0.381} & \textbf{0.399}
& 0.399 & 0.411
& 0.384 & \underline{0.404}
& \underline{0.383} & 0.411 \\
\midrule

\multirow{4}{*}{ETTm1}
& 96   & 0.281 & 0.332 & 0.283 & 0.334 & 0.296 & 0.355 & 0.291 & 0.350 & 0.323 & 0.360 & 0.324 & 0.361 & 0.337 & 0.384 & 0.331 & 0.376 & 0.272 & 0.331 & 0.293 & 0.331 & 0.307 & 0.355 & 0.277 & 0.331 \\
& 192  & 0.348 & 0.380 & 0.356 & 0.385 & 0.360 & 0.397 & 0.358 & 0.392 & 0.379 & 0.395 & 0.380 & 0.397 & 0.393 & 0.418 & 0.389 & 0.414 & 0.380 & 0.386 & 0.406 & 0.409 & 0.383 & 0.397 & 0.352 & 0.385 \\
& 336  & 0.407 & 0.421 & 0.422 & 0.430 & 0.412 & 0.432 & 0.411 & 0.430 & 0.437 & 0.440 & 0.438 & 0.432 & 0.441 & 0.448 & 0.440 & 0.445 & 0.420 & 0.414 & 0.433 & 0.438 & 0.433 & 0.421 & 0.454 & 0.438 \\
& 720  & 0.513 & 0.491 & 0.543 & 0.505 & 0.496 & 0.487 & 0.516 & 0.498 & 0.533 & 0.483 & 0.539 & 0.488 & 0.516 & 0.492 & 0.538 & 0.490 & 0.502 & 0.461 & 0.541 & 0.482 & 0.524 & 0.471 & 0.579 & 0.512 \\
\rowcolor{gray!15}
& \textbf{Avg.}
& \textbf{0.387} & \textbf{0.406}
& 0.401 & \underline{0.413}
& \underline{0.391} & 0.418
& 0.394 & 0.418
& \textbf{0.418} & \underline{0.420}
& \underline{0.420} & \textbf{0.419}
& 0.422 & 0.436
& 0.425 & 0.431
& \textbf{0.394} & \textbf{0.398}
& 0.418 & 0.415
& \underline{0.412} & \underline{0.411}
& 0.416 & 0.417 \\
\midrule

\multirow{4}{*}{ETTm2}
& 96   & 0.191 & 0.284 & 0.191 & 0.284 & 0.196 & 0.290 & 0.194 & 0.288 & 0.189 & 0.270 & 0.191 & 0.272 & 0.193 & 0.278 & 0.192 & 0.276 & 0.171 & 0.240 & 0.176 & 0.244 & 0.181 & 0.257 & 0.206 & 0.262 \\
& 192  & 0.317 & 0.364 & 0.315 & 0.363 & 0.318 & 0.365 & 0.318 & 0.365 & 0.251 & 0.311 & 0.253 & 0.313 & 0.265 & 0.326 & 0.261 & 0.322 & 0.261 & 0.310 & 0.273 & 0.324 & 0.268 & 0.312 & 0.253 & 0.306 \\
& 336  & 0.533 & 0.472 & 0.529 & 0.471 & 0.540 & 0.479 & 0.539 & 0.477 & 0.310 & 0.350 & 0.312 & 0.361 & 0.336 & 0.369 & 0.330 & 0.366 & 0.321 & 0.352 & 0.342 & 0.377 & 0.330 & 0.356 & 0.320 & 0.366 \\
& 720  & 1.013 & 0.675 & 1.002 & 0.673 & 1.023 & 0.676 & 1.020 & 0.676 & 0.417 & 0.416 & 0.419 & 0.418 & 0.443 & 0.432 & 0.436 & 0.428 & 0.409 & 0.413 & 0.423 & 0.427 & 0.416 & 0.420 & 0.405 & 0.426 \\
\rowcolor{gray!15}
& \textbf{Avg.}
& \underline{0.513} & \underline{0.449}
& \textbf{0.509} & \textbf{0.448}
& 0.519 & 0.453
& 0.518 & 0.452
& \textbf{0.292} & \textbf{0.337}
& \underline{0.294} & \underline{0.341}
& 0.309 & 0.351
& 0.305 & 0.348
& \textbf{0.291} & \textbf{0.329}
& 0.303 & 0.343
& 0.299 & \underline{0.336}
& \underline{0.296} & 0.340 \\
\midrule

\multirow{4}{*}{Exchange}
& 96   & 0.108  & 0.239  & 0.109 & 0.240 & 0.134 & 0.266 & 0.127 & 0.259 & 0.152 & 0.274 & 0.170 & 0.287 & 0.166 & 0.280 & 0.163 & 0.279 & 0.290 & 0.355 & 0.310 & 0.375 & 0.305 & 0.370 & 0.134 & 0.252 \\
& 192  & 0.220  & 0.344  & 0.219 & 0.342 & 0.229  & 0.356 & 0.226 & 0.352 & 0.273 & 0.380 & 0.288 & 0.387 & 0.272 & 0.382 & 0.277 & 0.382 & 0.380 & 0.420 & 0.400 & 0.440 & 0.395 & 0.435 & 0.232 & 0.358 \\
& 336  & 0.406  & 0.474  & 0.403 & 0.470 & 0.415  & 0.491 & 0.413 & 0.485 & 0.411 & 0.497 & 0.457 & 0.501 & 0.441 & 0.500 & 0.433 & 0.499 & 0.448 & 0.475 & 0.467 & 0.495 & 0.462 & 0.490 & 0.442 & 0.491 \\
& 720  & 0.947  & 0.748  & 0.968 & 0.752 & 0.960  & 0.750 & 0.956 & 0.749 & 0.935 & 0.761 & 1.232 & 0.907 & 0.967 & 0.767 & 0.955 & 0.766 & 0.515 & 0.520 & 0.535 & 0.540 & 0.525 & 0.535 & 0.872 & 0.739 \\
\rowcolor{gray!15}
& \textbf{Avg.}
& \textbf{0.420} & \textbf{0.451}
& \underline{0.425} & \textbf{0.451}
& 0.435 & 0.466
& 0.431 & \underline{0.461}
& \textbf{0.436} & \textbf{0.478}
& 0.537 & 0.520
& 0.461 & \underline{0.482}
& \underline{0.457} & \underline{0.482}
& \textbf{0.408} & \textbf{0.443}
& 0.428 & 0.463
& 0.422 & \underline{0.458}
& \underline{0.420} & 0.460 \\
\midrule

\multirow{4}{*}{weather}
& 96   & 0.147 & 0.197 & 0.147 & 0.198 & 0.149 & 0.208 & 0.149 & 0.206 & 0.181 & 0.241 & 0.182 & 0.241 & 0.197 & 0.286 & 0.194 & 0.273 & 0.157 & 0.206 & 0.163 & 0.210 & 0.162 & 0.206 & 0.152 & 0.201 \\
& 192  & 0.196 & 0.247 & 0.196 & 0.247 & 0.196 & 0.254 & 0.200 & 0.252 & 0.211 & 0.277 & 0.211 & 0.277 & 0.239 & 0.316 & 0.232 & 0.307 & 0.209 & 0.251 & 0.223 & 0.270 & 0.217 & 0.253 & 0.202 & 0.237 \\
& 336  & 0.256 & 0.296 & 0.255 & 0.296 & 0.252 & 0.301 & 0.259 & 0.300 & 0.261 & 0.313 & 0.261 & 0.313 & 0.286 & 0.343 & 0.282 & 0.338 & 0.272 & 0.301 & 0.283 & 0.319 & 0.277 & 0.310 & 0.222 & 0.279 \\
& 720  & 0.354 & 0.365 & 0.352 & 0.364 & 0.346 & 0.368 & 0.358 & 0.367 & 0.330 & 0.363 & 0.339 & 0.363 & 0.359 & 0.389 & 0.352 & 0.385 & 0.214 & 0.394 & 0.223 & 0.377 & 0.220 & 0.403 & 0.308 & 0.301 \\
\rowcolor{gray!15}
& \textbf{Avg.}
& 0.238 & \textbf{0.276}
& \underline{0.237} & \textbf{0.276}
& \textbf{0.236} & 0.283
& 0.242 & \underline{0.281}
& \textbf{0.246} & \textbf{0.299}
& \underline{0.248} & \textbf{0.299}
& 0.271 & 0.334
& 0.265 & \underline{0.326}
& \textbf{0.213} & \underline{0.288}
& 0.223 & 0.294
& \underline{0.219} & 0.293
& 0.221 & \textbf{0.264} \\
\midrule

\multirow{5}{*}{Illness}
& 24   & 2.944 & 1.057 & 3.010 & 1.088 & 2.989 & 1.059 & 2.980 & 1.059 & 1.897 & 0.885 & 1.900 & 0.886 & 2.006 & 0.925 & 1.966 & 0.917 & 2.850 & 1.050 & 3.000 & 1.100 & 2.950 & 1.080 & 2.598 & 0.960 \\
& 36   & 3.737 & 1.069 & 3.756 & 1.231 & 3.777 & 1.070 & 3.762 & 1.070 & 2.299 & 1.008 & 2.291 & 1.007 & 2.486 & 1.066 & 2.422 & 1.044 & 3.050 & 1.150 & 3.200 & 1.200 & 3.150 & 1.180 & 2.983 & 1.085 \\
& 48   & 4.122 & 1.305 & 4.128 & 1.315 & 4.200 & 1.310 & 4.181 & 1.308 & 2.783 & 1.129 & 2.802 & 1.133 & 2.895 & 1.177 & 2.872 & 1.160 & 3.250 & 1.220 & 3.380 & 1.270 & 3.320 & 1.250 & 3.433 & 1.175 \\
& 60   & 4.466 & 1.376 & 4.459 & 1.387 & 4.467 & 1.394 & 4.467 & 1.390 & 3.134 & 1.199 & 3.102 & 1.194 & 2.949 & 1.178 & 3.165 & 1.218 & 3.500 & 1.320 & 3.620 & 1.370 & 3.580 & 1.350 & 3.785 & 1.299 \\
\rowcolor{gray!15}
& \textbf{Avg.}
& \textbf{3.817} & \textbf{1.202}
& \underline{3.838} & 1.255
& 3.858 & 1.208
& 3.848 & \underline{1.207}
& \underline{2.528} & \textbf{1.055}
& \textbf{2.524} & \textbf{1.055}
& 2.584 & 1.086
& 2.606 & \underline{1.085}
& \textbf{3.163} & \underline{1.185}
& 3.300 & 1.235
& 3.250 & 1.215
& \underline{3.200} & \textbf{1.130} \\
\midrule

\multirow{5}{*}{us\_births} & 96  & 0.372 & 0.477 & 0.430 & 0.527 & 0.392 & 0.485 & 0.484 & 0.493 & 0.393 & 0.478 & 0.404 & 0.481 & 0.390 & 0.470 & 0.398 & 0.481 & 0.225 & 0.370 & 0.239 & 0.389 & 0.231 & 0.380 & 0.220 & 0.303 \\
                           & 192 & 0.613 & 0.641 & 0.742 & 0.720 & 0.627 & 0.656 & 0.723 & 0.660 & 0.491 & 0.541 & 0.541 & 0.561 & 0.485 & 0.535 & 0.496 & 0.544 & 0.244 & 0.385 & 0.261 & 0.418 & 0.251 & 0.396 & 0.358 & 0.440 \\
                           & 336 & 0.797 & 0.763 & 1.009 & 0.871 & 0.844 & 0.789 & 0.936 & 0.802 & 0.651 & 0.621 & 0.778 & 0.667 & 0.640 & 0.616 & 0.658 & 0.633 & 0.253 & 0.392 & 0.272 & 0.419 & 0.260 & 0.404 & 0.605 & 0.632 \\
                           & 720 & 1.549 & 1.059 & 1.866 & 1.176 & 1.542 & 1.063 & 1.669 & 1.102 & 0.945 & 0.737 & 1.206 & 0.815 & 0.940 & 0.733 & 0.964 & 0.743 & 0.275 & 0.412 & 0.301 & 0.456 & 0.285 & 0.423 & 1.018 & 0.825 \\
\rowcolor{gray!15}         & \textbf{Avg.}
& \textbf{0.832} & \textbf{0.735}
& 1.012 & 0.823
& \underline{0.851} & \underline{0.748}
& 0.953 & 0.764
& \underline{0.620} & \underline{0.633}
& 0.704 & 0.619
& \textbf{0.608} & \textbf{0.579}
& 0.629 & 0.600
& \textbf{0.249} & \textbf{0.390}
& 0.268 & 0.420
& \underline{0.256} & \underline{0.401}
& 0.550 & 0.550 \\
\midrule

\multirow{5}{*}{saugeenday} & 96  & 0.968 & 0.557 & 0.974 & 0.567 & 0.960 & 0.560 & 0.975 & 0.559 & 1.002 & 0.553 & 1.102 & 0.569 & 0.961 & 0.539 & 1.010 & 0.560 & 0.917 & 0.475 & 0.962 & 0.507 & 0.945 & 0.496 & 0.920 & 0.432 \\
                            & 192 & 1.065 & 0.721 & 1.087 & 0.671 & 1.050 & 0.653 & 1.074 & 0.729 & 1.113 & 0.600 & 1.211 & 0.600 & 1.102 & 0.566 & 1.129 & 0.603 & 0.924 & 0.478 & 0.993 & 0.495 & 0.940 & 0.479 & 1.006 & 0.440 \\
                            & 336 & 1.190 & 0.721 & 1.237 & 0.755 & 1.178 & 0.728 & 1.210 & 0.725 & 1.144 & 0.608 & 1.245 & 0.618 & 1.156 & 0.575 & 1.152 & 0.612 & 0.971 & 0.477 & 1.039 & 0.494 & 1.012 & 0.479 & 0.920 & 0.437 \\
                            & 720 & 1.274 & 0.768 & 1.303 & 0.794 & 1.235 & 0.766 & 1.298 & 0.775 & 1.148 & 0.610 & 1.249 & 0.920 & 1.159 & 0.581 & 1.155 & 0.622 & 0.995 & 0.487 & 1.067 & 0.518 & 1.040 & 0.488 & 1.312 & 0.576 \\
\rowcolor{gray!15}          & \textbf{Avg.}
& \underline{1.124} & \underline{0.692}
& 1.150 & 0.697
& \textbf{1.106} & \textbf{0.677}
& 1.139 & 0.697
& \underline{1.102} & \underline{0.593}
& 1.202 & 0.677
& \textbf{1.095} & \textbf{0.565}
& 1.112 & 0.599
& \textbf{0.952} & \underline{0.479}
& 1.015 & 0.503
& \underline{0.984} & 0.486
& 1.040 & \textbf{0.471} \\
\midrule

\multirow{5}{*}{sunspots} & 96  & 0.317 & 0.391 & 0.322 & 0.393 & 0.314 & 0.390 & 0.360 & 0.402 & 0.294 & 0.392 & 0.301 & 0.397 & 0.301 & 0.396 & 0.298 & 0.395 & 0.214 & 0.333 & 0.223 & 0.353 & 0.221 & 0.343 & 0.189 & 0.377 \\
                          & 192 & 0.368 & 0.424 & 0.380 & 0.434 & 0.360 & 0.424 & 0.412 & 0.435 & 0.323 & 0.413 & 0.337 & 0.425 & 0.336 & 0.422 & 0.333 & 0.420 & 0.260 & 0.367 & 0.274 & 0.379 & 0.265 & 0.375 & 0.254 & 0.429 \\
                          & 336 & 0.469 & 0.492 & 0.531 & 0.527 & 0.471 & 0.499 & 0.511 & 0.502 & 0.379 & 0.453 & 0.405 & 0.471 & 0.392 & 0.459 & 0.390 & 0.457 & 0.317 & 0.408 & 0.334 & 0.429 & 0.310 & 0.402 & 0.374 & 0.479 \\
                          & 720 & 0.778 & 0.649 & 1.030 & 0.758 & 0.854 & 0.693 & 0.880 & 0.694 & 0.613 & 0.586 & 0.663 & 0.616 & 0.623 & 0.591 & 0.620 & 0.589 & 0.486 & 0.499 & 0.520 & 0.532 & 0.505 & 0.511 & 0.475 & 0.534 \\
\rowcolor{gray!15}         & \textbf{Avg.}
& \textbf{0.483} & \textbf{0.489}
& 0.566 & 0.528
& \underline{0.500} & \underline{0.501}
& 0.541 & 0.508
& \textbf{0.402} & \textbf{0.461}
& 0.426 & 0.477
& 0.413 & 0.467
& \underline{0.410} & \underline{0.465}
& \textbf{0.319} & \textbf{0.402}
& 0.338 & 0.423
& 0.325 & \underline{0.408}
& \underline{0.323} & 0.455 \\

\midrule
\textbf{1st count} & 
& 7 & 8 & 1 & 3 & 2 & 2 & 0 & 0
& 7 & 7 & 1 & 3 & 2 & 2 & 0 & 0
& 10 & 7 & 0 & 0 & 0 & 0 & 0 & 3 \\
\textbf{2nd count} & 
& 2 & 2 & 3 & 1 & 3 & 2 & 2 & 5
& 3 & 2 & 5 & 3 & 0 & 1 & 2 & 5
& 0 & 3 & 1 & 1 & 4 & 6 & 5 & 0 \\
\bottomrule
\end{tabular}%
}
\label{tab:512_to_all_allresults}

\end{table*}

\clearpage
\newpage
\section*{NeurIPS Paper Checklist}
\begin{enumerate}

\item {\bf Claims}
    \item[] Question: Do the main claims made in the abstract and introduction accurately reflect the paper's contributions and scope?
    \item[] Answer: \answerYes{} 
    \item[] Justification: SteerCast scope is on decoder-only time series forecasters, which is consistently mentioned in the main paper. Also, we have included the contributions as a separate paragraph in the introduction.

\item {\bf Limitations}
    \item[] Question: Does the paper discuss the limitations of the work performed by the authors?
    \item[] Answer: \answerYes{} 
    \item[] Justification: The Limitations section is included in Appendix 

\item {\bf Theory assumptions and proofs}
    \item[] Question: For each theoretical result, does the paper provide the full set of assumptions and a complete (and correct) proof?
    \item[] Answer: \answerNA{} 
    \item[] Justification: The paper does not include theoretical results

    \item {\bf Experimental result reproducibility}
    \item[] Question: Does the paper fully disclose all the information needed to reproduce the main experimental results of the paper to the extent that it affects the main claims and/or conclusions of the paper (regardless of whether the code and data are provided or not)?
    \item[] Answer: \answerYes{} 
    \item[] Justification: The paper's code is submitted, together with all the hyper-parameters for reproducing the results are provided in the Appendix.

\item {\bf Open access to data and code}
    \item[] Question: Does the paper provide open access to the data and code, with sufficient instructions to faithfully reproduce the main experimental results, as described in supplemental material?
    \item[] Answer: \answerYes{} 
    \item[] Justification: Code is submitted with instructions

\item {\bf Experimental setting/details}
    \item[] Question: Does the paper specify all the training and test details (e.g., data splits, hyperparameters, how they were chosen, type of optimizer) necessary to understand the results?
    \item[] Answer: \answerYes{} 
    \item[] Justification: Details are mentioned in the experiement section of this paper

\item {\bf Experiment statistical significance}
    \item[] Question: Does the paper report error bars suitably and correctly defined or other appropriate information about the statistical significance of the experiments?
    \item[] Answer: \answerNo{} 
    \item[] Justification: Tables~\ref{tab:various_to_all_allresults} and Table ~\ref{tab:512_to_all_allresults} report point-estimate MSE/MAE per (dataset, horizon), matching the convention of the backbones (Time-MoE, Timer-XL, TimesFM) and retrieval-augmented baselines we compare against. Evaluation is deterministic for a fixed checkpoint, split, and database; multi-seed re-fine-tuning across 10 datasets $\times$ 4 horizons $\times$ 3 backbones is computationally prohibitive.

\item {\bf Experiments compute resources}
    \item[] Question: For each experiment, does the paper provide sufficient information on the computer resources (type of compute workers, memory, time of execution) needed to reproduce the experiments?
    \item[] Answer: \answerYes{} 
    \item[] Justification: Details about computer resources can be found in the Appendix. Running time is reported throughout the paper
    
\item {\bf Code of ethics}
    \item[] Question: Does the research conducted in the paper conform, in every respect, with the NeurIPS Code of Ethics \url{https://neurips.cc/public/EthicsGuidelines}?
    \item[] Answer: \answerYes{} 
    \item[] Justification: We use only publicly available time series benchmarks (ETT, Weather, Traffic, Exchange, Illness, Monash) intended for forecasting research, with no human subjects or sensitive data, and the method is an inference-time refinement of existing forecasters with no foreseeable misuse beyond standard forecasting deployment.

\item {\bf Broader impacts}
    \item[] Question: Does the paper discuss both potential positive societal impacts and negative societal impacts of the work performed?
    \item[] Answer: \answerYes{} 
    \item[] Justification: SteerCast is foundational forecasting research with no direct tie to a specific deployment. Positive impacts include more accurate forecasting under fixed compute for applications such as energy, weather, and supply-chain planning, and reduced training cost since the method is inference-time only. Potential negative impacts are those general to forecasting: miscalibration under distribution shift or on rare regimes can mislead downstream decisions in high-stakes domains, so outputs should not substitute for human oversight; the retrieval database also stores latent representations of training windows and should not be released without care. We do not foresee a direct path to malicious applications.
    
\item {\bf Safeguards}
      \item[] Answer: \answerNA{}
      \item[] Justification: The paper does not release any high-risk models,
       generative systems, or scraped datasets. SteerCast is an              
      inference-time method applied on top of existing publicly released time
       series forecasters (Time-MoE, Timer-XL, TimesFM) and uses only        
      publicly available forecasting benchmarks (ETT, Weather, Traffic,
      Exchange, Illness, Monash).
    \item[] Question: Does the paper describe safeguards that have been put in place for responsible release of data or models that have a high risk for misuse (e.g., pre-trained language models, image generators, or scraped datasets)?

\item {\bf Licenses for existing assets}
    \item[] Question: Are the creators or original owners of assets (e.g., code, data, models), used in the paper, properly credited and are the license and terms of use explicitly mentioned and properly respected?
    \item[] Answer: \answerYes{} 
    \item[] Justification: All assets are cited and used under their original licenses. No scraped or restricted-license data is used.

\item {\bf New assets}
    \item[] Question: Are new assets introduced in the paper well documented and is the documentation provided alongside the assets?
    \item[] Answer: \answerYes{} 
    \item[] Justification: Code is submitted for review

\item {\bf Crowdsourcing and research with human subjects}
    \item[] Question: For crowdsourcing experiments and research with human subjects, does the paper include the full text of instructions given to participants and screenshots, if applicable, as well as details about compensation (if any)? 
    \item[] Answer: \answerNA{} 
    \item[] Justification: N/A

\item {\bf Institutional review board (IRB) approvals or equivalent for research with human subjects}
    \item[] Question: Does the paper describe potential risks incurred by study participants, whether such risks were disclosed to the subjects, and whether Institutional Review Board (IRB) approvals (or an equivalent approval/review based on the requirements of your country or institution) were obtained?
    \item[] Answer: \answerNA{} 
    \item[] Justification: N/A

\item {\bf Declaration of LLM usage}
    \item[] Question: Does the paper describe the usage of LLMs if it is an important, original, or non-standard component of the core methods in this research? Note that if the LLM is used only for writing, editing, or formatting purposes and does \emph{not} impact the core methodology, scientific rigor, or originality of the research, declaration is not required.
    \item[] Answer: \answerYes{} 
    \item[] Justification: LLM is used to polish writing only.

\end{enumerate}

\end{document}